ORIGINAL ARTICLE

# Equipment-centric workpiece localization in near real-time using deep learning-based vision and event-driven finite state machines

**Dohyeon Kong[1] · Jaebong Cho[1] · Hyunbo Cho[1]**

## Abstract

Continuous workpiece localization is critical for ensuring traceability, quality control, and process coordination in hot forging environments, where heavy workpieces are repeatedly transferred by specialized handling equipment such as overhead cranes among storage yards, furnaces, and forging stations. However, traditional workpiece-centric approaches are often unreliable under extreme temperatures, surface degradation, and irregular routing conditions, leaving a critical gap in continuous localization. To overcome these challenges, this study presents an equipment-centric framework that infers workpiece locations indirectly by analyzing equipment operations captured through video streams from multiple static 2D cameras. This paradigm shift from workpiece-centric to equipment-centric localization enables robust inference under harsh conditions by coupling equipment behavior with workpiece-handling events. The framework estimates floorplan-space 3D coordinates of handling equipment and recognizes workpiece-handling activities such as grasp and release. Event-driven finite state machines (FSMs) validate and detect discrete handling events based on equipment coordinates and recognized activity cues, enabling continuous inference and updating of workpiece states and floorplan-space coordinates. Experimental validation in an operational hot forging factory demonstrated the effectiveness and practicality of the proposed framework, achieving 100 % event detection accuracy within a 33-second tolerance window, a mean localization error of 317.8 mm, and a mean system latency of 21 seconds. Ablation studies further confirmed that integrating the Keypoint-Guided Attention (KPGA) mechanism improves activity recognition performance compared to baseline 3D convolutional neural networks (3D-CNNs) and transformer-based models. Beyond localization accuracy, the structured outputs of the framework bridge vision-based perception and event-driven reasoning by representing handling operations as interpretable state transitions. This capability enables data-driven visualization of workpiece transfers and quantitative evaluation of equipment handling states, contributing to more intelligent and traceable forging operations.

## 1 Introduction

Metal forming processes, particularly forging, are essential for producing high-strength components used in industries such as automotive, aerospace, and heavy machinery. Hot forging, which involves shaping metal above its recrystallization temperature, enables deformation without cracking and refines the workpiece's grain structure. Such processes require transfer of workpieces between furnaces, forging presses, and storage yards. Workpieces can weigh hundreds of tons and are moved using specialized handling equipment such as overhead cranes [1]. Such processes require transfer of workpieces between furnaces, forging presses, and storage yards.

✉ Hyunbo Cho
hcho@postech.ac.kr

Dohyeon Kong
dohyeono@postech.ac.kr

Jaebong Cho
jaycho00@postech.ac.kr

[1] Department of Industrial and Management Engineering, Pohang University of Science and Technology (POSTECH), Pohang 37673, Republic of Korea

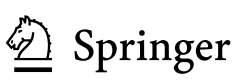

Continuous localization of workpieces is a fundamental requirement in hot forging, as it enables timely identification and location awareness throughout the production lifecycle. It serves as the foundation for effective tracking, which involves monitoring the status and progress of each workpiece across various processing steps. In modern production paradigms such as lot-size-one manufacturing, where each workpiece is assigned a unique routing tailored to specific customer requirements, the importance of accurate and continuous localization becomes even more pronounced [2]. Errors in localization can lead to tracking errors such as misidentification or misplacement of workpieces. These tracking errors, in turn, may result in skipped forging steps, incorrect heat treatments, or deviation from required specifications, ultimately compromising product quality and customer satisfaction. Robust continuous localization is therefore essential for enabling accurate tracking in individualized, high-mix manufacturing environments.

Despite this need, localization in hot forging environments remains challenging due to extreme operating conditions. Factors such as high temperatures, forging pressure, and surface contamination make it virtually impossible to attach physical tags such as radio frequency identification (RFID) and ultra-wideband (UWB) to the workpiece. These tags are essential components of tag-based localization systems, which are the most predominant and widely adopted approach in general industrial settings [3]. In response to these limitations, existing studies in hot forging environments have explored alternative localization strategies that do not rely on conventional physical tags. These approaches can be broadly categorized into identifier-based methods, which use durable markings on the workpiece combined with vision-based recognition, and identifier-free methods, which leverage the inherent visual or geometric features of the workpiece.

Zhao et al. [4] proposed a character marking strategy combined with computer vision-based recognition. This method involved stamping visible identifiers onto the workpiece surface, offering an alternative to physical tags. However, its recognition accuracy is affected by surface contamination and variations in lighting conditions. To enhance the durability of identifiers, Liewald et al. [5] investigated mechanically robust markings, including needled stamps and laser-engraved quick response (QR) codes applied to the surfaces. These identifiers were experimentally shown to withstand high temperatures and mechanical wear, making them suitable for the forging environments. Nonetheless, successful recognition still depends on proper workpiece orientation and visibility within the camera's field of view, which limits their effectiveness in dynamic production settings. Moving beyond physical markings, Kang et al. [6] proposed a deep learning-based framework that identifies workpieces by analyzing their surface texture patterns. This non-intrusive approach avoids the use of the identifiers by leveraging the natural appearance of the workpiece. However, its performance is highly sensitive to evolving surface conditions. As textures change over time due to oxidation and deformation, the approach requires periodic retraining of the deep learning models to maintain reliable identification accuracy.

While these approaches have advanced workpiece localization in hot forging environments, they share a fundamental limitation rooted in their reliance on direct observation of the workpiece for identification. This restricts localization to fixed checkpoints and prevents continuous tracking. The limitation is critical in hot forging, where process routings are irregular. Workpieces are not transferred along fixed routes or at regular intervals but are moved manually or semi-automatically based on operator judgment and production needs. They are often transferred arbitrarily within large workspaces such as storage yards.

A promising and underexplored direction for workpiece localization in hot forging environments is to shift from a workpiece-centric to an equipment-centric perspective. Since every change in a workpiece's location is inherently linked to equipment operations, monitoring equipment activities provides a scalable, non-intrusive proxy for inferring workpiece locations. Leveraging this insight, we aim to achieve near real-time, continuous localization without relying on physical tags, conventional identifiers, or direct visual access to the workpiece.

To realize this concept, we propose an equipment-centric framework that infers workpiece locations using video streams from multiple static 2D cameras. The proposed framework was implemented in a real-world hot forging factory comprising a yard, five furnaces, and a forging station, where two overhead cranes handle the transfer of heavy workpieces among stations. The framework simultaneously estimates floorplan-space 3D coordinates of equipment and recognizes workpiece-handling activities such as grasp and release, which are then combined through an event-driven finite state machine (FSM) to validate the activities as discrete events and logically infer and update the workpiece states and coordinates over time. In addition, the framework generates structured outputs that serve as a foundation for further analysis, including visualization of workpiece transfers and process stages, as well as analysis of equipment utilization—supporting process optimization, workload balancing, and data-driven management in hot forging environments. The main contributions are as follows:

- An equipment-centric framework for continuous workpiece localization based on estimated floorplan-space 3D coordinates and recognized handling activities of equipment.

- A multi-view 3D localization pipeline that reconstructs equipment trajectories as floorplan-space 3D coordinates through pose estimation and spatial association with the factory floorplan.
- A keypoint-guided attention (KPGA) mechanism integrated into a 3D convolutional neural network (3D-CNN) to enhance activity recognition by leveraging structural information of the equipment.
- An event-driven finite state machine (FSM) that validates recognized activities as discrete events and logically infers and updates continuous workpiece states and coordinates.
- Real-world deployment and evaluation of the proposed framework in an operational hot forging environment, demonstrating its effectiveness for both localization and operational analysis.

## 2 Related works

This section reviews prior studies on vision-based localization and equipment activity recognition, which jointly support continuous, equipment-centric workpiece localization.

### 2.1 Vision-based localization

Vision-based localization has emerged as an effective alternative to traditional tag-based systems, particularly in industrial environments where attaching physical tags such as RFID or UWB is impractical. Its primary objective is to enable continuous and non-intrusive localization of objects through visual observations. One of the key advantages of vision-based methods is their minimal infrastructure requirement, often leveraging existing camera systems such as surveillance setups. These methods eliminate the need for physical interaction with objects and integrate seamlessly into existing operational workflows. Furthermore, vision-based localization provides rich contextual information, including movement patterns and object interactions, making it particularly suitable for extreme industrial environments like hot forging, where robustness and low maintenance are essential.

To systematically categorize existing vision-based localization methods, Morar et al. [7] proposed a taxonomy based on three main elements: environment data, sensing devices, and detected elements. Environment data includes prior knowledge of the monitored space, such as known camera placements, factory floorplans, or stored reference images. Sensing devices refer to the camera configurations used for data collection, distinguishing between fixed (static) and mobile cameras, as well as 2D or 3D camera technologies. Detected elements describe the visual features used for localization, ranging from artificial markers like QR codes to natural, real-world features such as machinery, personnel, or structural elements. The current study focuses specifically on approaches employing static 2D cameras installed, utilizing real environmental features. This configuration is particularly well suited to hot forging environments. Static cameras can be safely installed in overhead, thermally shielded positions and maintained with minimal interference to operations. Compared to depth or mobile sensing devices, 2D cameras are cost-effective, robust, and often already present as part of existing surveillance infrastructure. Moreover, detecting real features such as cranes, floor structures, or equipment geometry avoids the need for physical tags, which are impractical due to extreme heat and surface contamination.

Early vision-based localization methods commonly relied on conventional image-processing techniques and were typically deployed in structured indoor settings, such as offices, retail stores, and public areas. For instance, Shim and Cho [8] applied background subtraction and motion detection using overhead cameras in indoor public spaces. They mapped detected object locations onto a 2D floorplan through homography transformations calibrated with known spatial reference points. Similarly, Dias and Jorge [9] utilized multiple static cameras across office spaces, employing blob tracking following background subtraction. Detected objects were localized onto a unified spatial coordinate system using homographic mapping, enabling seamless tracking across multiple camera views. Sun et al. [10] used a panoramic ceiling-mounted camera for indoor surveillance. They extracted foreground regions via frame differencing and projected detections onto the floorplan using a radial projection model based on camera height and central position. Although these traditional methods achieved reliable results under controlled, stable conditions, they exhibited notable limitations when faced with complex environments, including sensitivity to lighting variations, shadows, occlusions, and background clutter. Additionally, the manual tuning and precise calibration requirements restricted their adaptability to the extreme industrial environments.

To overcome these limitations, recent approaches have integrated artificial intelligence (AI), especially deep learning methods, significantly enhancing localization robustness. AI-driven methods have been applied effectively in more complex scenarios, including smart buildings, warehouses, transportation hubs, and construction sites, where visual complexity and occlusion are common. For instance, Cosma et al. [11] developed a pedestrian localization system using a pose estimation network to detect skeletal keypoints. These keypoints, particularly foot positions, were mapped onto a 2D floorplan using perspective transformations, leveraging pre-calibrated camera geometry. Jain et al. [12] addressed mobile user localization by employing upward-facing smartphone cameras. Their approach used CNN-based object detection to identify ceiling landmarks, such as lights and vents, matching them with known positions on the 2D floorplans for accurate localization. Morar et al. [7] further advanced AI-based localization in a smart workspace, combining instance segmentation (Mask R-CNN) and human pose estimation (OpenPose) to detect multiple individuals. They subsequently mapped the identified body keypoints onto the 2D floorplan using homographic transformations, facilitating real-time localization enriched with contextual understanding. Pfitzner et al. [13] expanded real-time localization into complex construction environments, tracking workers, vehicles, and materials using multi-camera object detection integrated into building information modeling (BIM)-based layout using perspective transformations, improving accuracy across expansive sites. However, challenges remained in accurately localizing objects positioned at elevated heights, such as cranes or structural pillars.

A critical research gap persists in vision-based localization, particularly concerning environments involving mid-air object locations. Traditional methods that rely on perspective-based geometric transformations typically assume movement occurs on a fixed 2D plane, a condition frequently violated in dynamic industrial contexts like hot forging factories, where equipment routinely moves vertically. Our study addresses this limitation by integrating a pose estimation network with multi-view perspective transformations to achieve accurate 3D localization within the factory floorplan.

## 2.2 Vision-based equipment activity recognition

Vision-based equipment activity recognition utilizes computer vision and deep learning technologies to automatically detect, interpret, and classify industrial equipment activities. By analyzing video data, these systems can recognize specific equipment activities, monitor operation patterns, and assess overall productivity. Such capabilities are integral to modern industrial automation, providing critical insights for effective equipment management, workflow optimization, and operational efficiency assessments.

A fundamental challenge in vision-based activity recognition is effectively capturing spatiotemporal features – that is, simultaneously understanding spatial relationships within frames and the temporal continuity across frame sequences. Without clear temporal modeling, systems struggle to differentiate between visually similar but semantically different activities, such as an overhead crane grasping versus releasing a workpiece, actions that share overlapping spatial appearances but differ fundamentally in motion dynamics. Recent advances in extracting meaningful spatiotemporal information have significantly improved recognition accuracy and reliability, enabling precise and efficient industrial monitoring.

Kim et al. [14] introduced a spatiotemporal reasoning framework specifically for excavator activity recognition, using object detection combined with heuristic rules. Their approach relied on the tracking-learning-detection (TLD) algorithm and frame differencing techniques to identify changes in equipment frame-space coordinates and orientation. Although spatial reasoning improved classification accuracy, their method lacked comprehensive temporal modeling, treating each frame independently, thus failing to effectively capture longer action sequences and overlapping activities. This highlighted the need for advanced, deep-learning-based temporal modeling.

Addressing the temporal modeling limitations, Kim and Chi [15] developed a hybrid model combining convolutional neural networks (CNNs) and long short-term memory (LSTM) networks, specifically designed to capture temporal dependencies in equipment workflows. Their system integrated CNNs for spatial feature extraction with double-layer LSTM (DLSTM) units for temporal analysis, enabling recognition outcomes that closely align with typical activity sequences (e.g., excavator actions such as digging, hauling, and dumping). This multi-stage temporal modeling approach notably improved robustness and accuracy, particularly in distinguishing closely timed and visually similar actions.

Roberts and Golparvar-Fard [16] further tackled activity recognition challenges in highly dynamic construction settings characterized by occlusion and clutter. Their method combined CNN-based object detection with probabilistic sequential modeling using Hidden Markov Models (HMMs) and Gaussian Mixture Models (GMMs). The object detector detected construction equipment across frames, while HMMs modeled sequential activity transitions probabilistically. Although this approach improved stability under noisy conditions, it still relied heavily on probability-driven state transitions, limiting the ability to directly learn complex motion patterns inherent in construction operations.

Chen et al. [17] emphasized detailed activity classification over extended sequences in excavator monitoring. Their system combined Faster Region-Based CNN (Faster R-CNN) for precise excavator detection, Deep Simple Online and Realtime Tracking (Deep SORT) for object tracking, and a 3D Residual Network (3D ResNet) to capture detailed spatiotemporal features. This enabled continuous recognition of complex activities over extended durations, significantly improving classification accuracy by effectively filtering out irrelevant motion.

Wang et al. [18] proposed a keypoint-based framework to address common industrial activity recognition issues, including occlusion and ambiguous activity states. Their approach utilized Keypoint-RCNN to detect critical mechanical keypoints, which were subsequently processed through a Heterogeneous Graph Convolutional Network (HGCN) to analyze spatial relationships among keypoints. While effective for enhancing spatial understanding, their single-frame focus limited comprehensive temporal analysis, highlighting an ongoing need for improved temporal modeling.

Despite significant advancements, conventional activity recognition methods typically analyze entire video frames without considering the importance of domain-specific knowledge or localized activity regions. Attention mechanisms have recently emerged as a solution to dynamically enhance relevant features. For instance, Li et al. [19] developed the Nesting Spatiotemporal Attention (NST) network, refining both spatial and temporal feature extraction within 3D-CNNs. Building upon these advances, our study proposes integrating a keypoint-guided attention mechanism, explicitly focusing on areas around keypoints where relevant activities occur, thereby significantly enhancing the accuracy and interpretability of vision-based equipment activity recognition.

## 3 Proposed framework

As illustrated in Fig. 1, the proposed framework integrates two stages. The equipment tracking stage estimates floor-plan-space coordinates and handling activities through pose estimation, floor localization, and activity classification. The workpiece localization stage interprets these outputs using event-driven FSMs to update states and coordinates of workpiece. The following subsections explain each stage in detail.

### 3.1 Data assumptions

The proposed framework operates on synchronized multi-view video data captured from static cameras $\mathcal{C}$. The camera set is partitioned into longitudinal views $\mathcal{C}_{long}$, aligned with the primary axis of the rectangular floorplan, and transverse views $\mathcal{C}_{trans}$, aligned with the perpendicular axis. To ensure geometric diversity for 3D localization, each subset must contain at least one camera, i.e., $\mathcal{C} = \mathcal{C}_{long} \cup \mathcal{C}_{trans}$ with $|\mathcal{C}_{long}| \geq 1$ and $|\mathcal{C}_{trans}| \geq 1$.

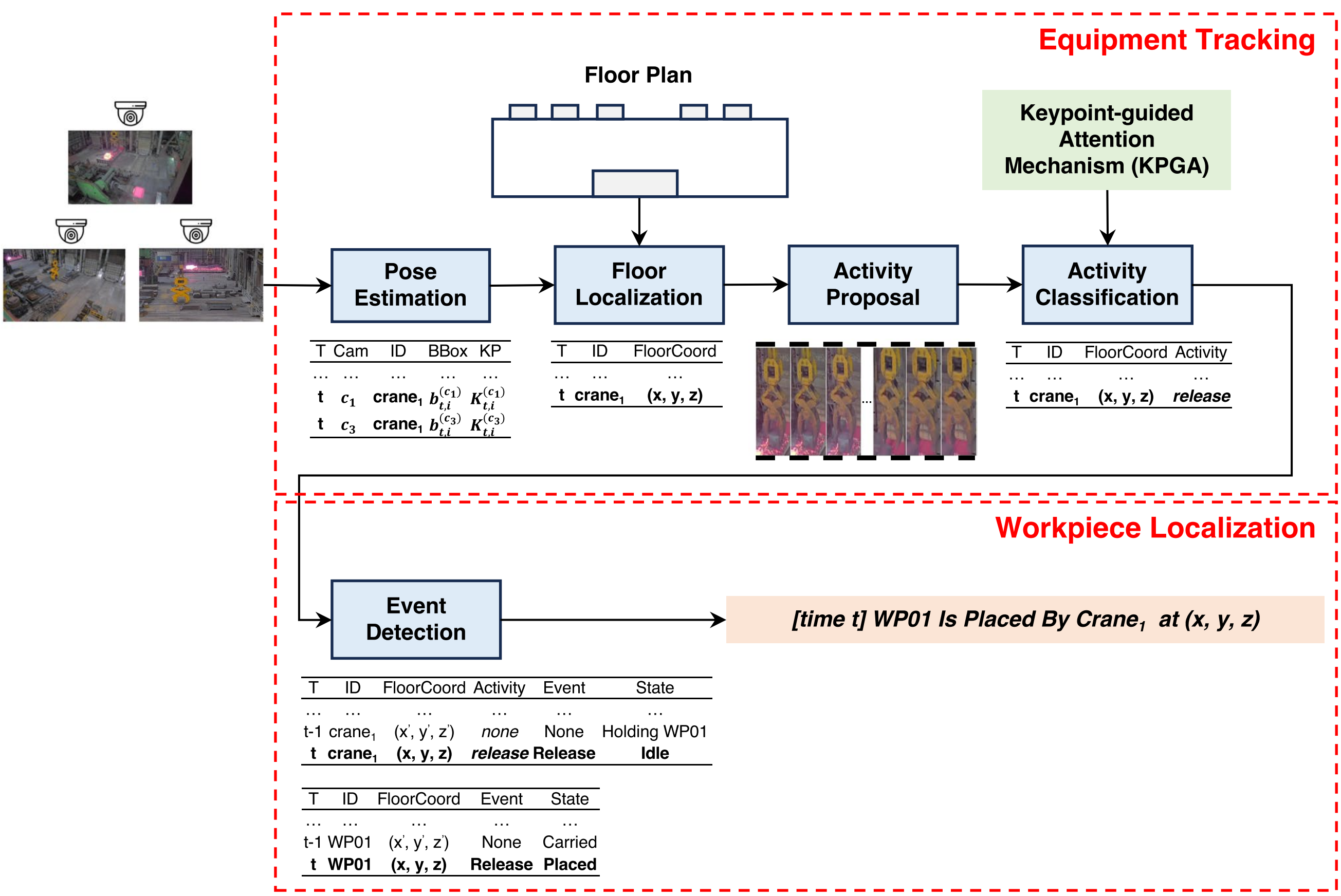

**Fig. 1** An overview of the proposed framework including equipment tracking and workpiece localization stages

At each time $t$, camera $c \in \mathcal{C}$ produces a frame $F_t^{(c)} \in \mathbb{R}^{H \times W \times 3}$, where $H$ and $W$ denote spatial resolution and the three channels correspond to RGB values. Collectively, $\{F_t^{(c)}\}_{c \in \mathcal{C}}$ form the synchronized multi-view input on which the subsequent stages of the framework operate.

## 3.2 Equipment tracking

### 3.2.1 Equipment location tracking

This stage continuously localizes equipment by transforming frame-space detections into 3D coordinates within floorplan-space. The process consists of pose estimation and floor localization.

**Pose estimation** The pose estimation module analyzes each frame $F_t^{(c)}$ and produces detections,

$$\mathrm{PE}(F_t^{(c)}) = O_t^{(c)} = \{o_{t,i}^{(c)}\}_{i=1}^{N_t} \tag{1}$$

where $N_t$ is the number of equipment instances detected. Each detection $o_{t,i}^{(c)}$ consists of a class label $\mathrm{cls}_{t,i}^{(c)}$, bounding box $b_{t,i}^{(c)}$, and keypoints $K_{t,i}^{(c)} = \{k_{t,i}^{(c,m)} \in \mathbb{R}^2 \mid m = 1, \ldots, M_{t,i}^{(c)}\}$.

Keypoints capture functionally meaningful parts such as crane grippers or lifting hooks, providing spatial and semantic cues beyond bounding boxes. Since the equipment set $E$ is finite and fixed, a simple object tracker associates detections across frames to preserve identity over time. From these associations, a representative point $p_{t,e}^{(c)}$ is derived as a linear combination of keypoints, capturing the functionally central region of each equipment.

**Floor localization** The floor localization module projects frame-space detections into floorplan-space 3D coordinates (Fig. 2). For equipment $e$, its state is represented as $(l_{t,e}, s_{t,e})$, where $l_{t,e}$ is the categorical region label and $s_{t,e} = (x_{t,e}, y_{t,e}, z_{t,e})$ are the estimated 3D coordinates.

Each camera $c$ is associated with a homography matrix $H_c \in \mathbb{R}^{3\times3}$, which maps frame-space coordinates to the floorplan-space $xy$-plane under the assumption of a reference height $z = 0$. However, when equipment is elevated above the floor, direct homography projection introduces

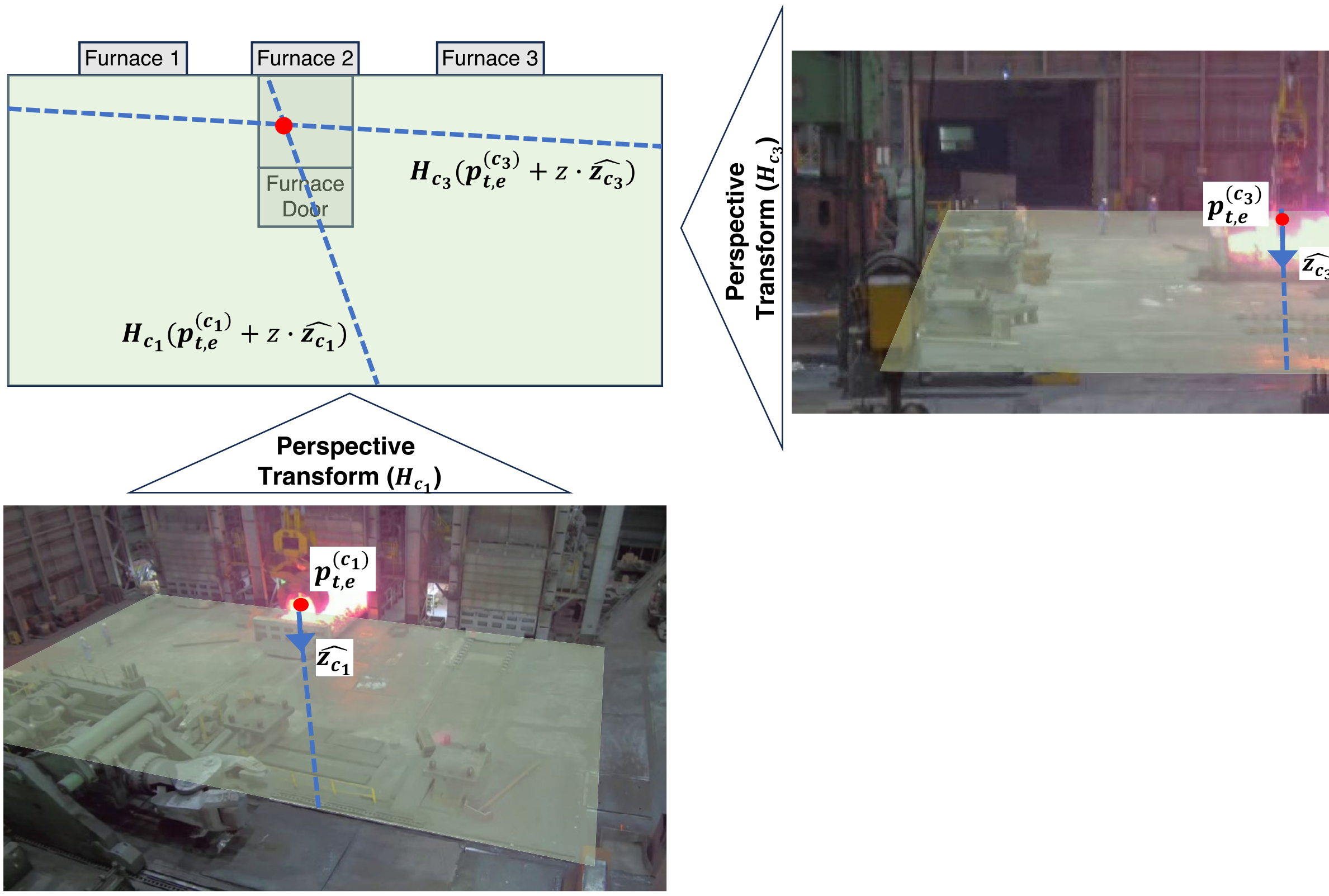

**Fig. 2** Estimation of floorplan-space equipment coordinates from multi-view camera detections using homography projections and height-dependent displacement corrections

systematic deviations due to geometric distortions from perspective effects [20]. To correct for these distortions, each camera is further assigned a predefined vertical displacement vector $\hat{z}_c \in \mathbb{R}^2$, which models the planar shift in projected coordinates as a function of increasing height.

The adjusted projection of a representative point $p_{t,e}^{(c)}$ at height $z$ is therefore computed as

$$q_{t,e}^{(c)}(z) = H_c\left(p_{t,e}^{(c)} + z \cdot \hat{z}_c\right). \tag{2}$$

The equipment height $z_{t,e}$ is then estimated by minimizing the discrepancy among projections across multiple views:

$$z_{t,e} = \arg\min_{z} \min_{(x,y)} \sum_{c \in \mathcal{C}'} \left\| q_{t,e}^{(c)}(z) - (x,y) \right\|^2, \tag{3}$$

where $\mathcal{C}' \subseteq \mathcal{C}$ includes at least one longitudinal and one transverse camera to ensure reliable depth estimation.
Finally, the $xy$-coordinates of the equipment are obtained as the mean of adjusted projections at the estimated height, and the categorical region label $l_{t,e}$ is determined by testing whether these coordinates fall within predefined spatial boundaries on the floorplan.

### 3.2.2 Equipment activity tracking

This stage aims to recognize equipment activities related to workpiece handling. The equipment activity proposal identifies candidate video clips likely containing relevant activities by detecting intervals where the equipment remains stationary. These clips serve as input for subsequent classification. The equipment activity classification then assigns each clip a semantic label – *grasp*, *release*, or *none* – indicating whether the equipment is engaged in a handling activity involving a workpiece.

**Activity proposal** An activity proposal is defined as a continuous time interval during which equipment displacement on the floorplan remains below a predefined threshold $\theta$. This criterion is based on the assumption that handling activities occur while the equipment is largely stationary. Formally, given localization results $(l_{t,e}, s_{t,e})$ for equipment $e \in E' \subset E$, where $E'$ denotes the subset of equipment capable of handling workpieces (e.g., cranes), an activity proposal interval $(t_{\text{start}}, t_{\text{end}})$ satisfies the condition defined as follows:

$$\|s_{t',e} - s_{t'+1,e}\| \leq \theta, \quad \forall t' \in [t_{\text{start}}, t_{\text{end}} - 1] \tag{4}$$

The proposal is finalized at time $t = t_{\text{end}} + 1$, when the displacement condition is violated. At this point, the corresponding video clip $V^e_{(t_{\text{start}}, t_{\text{end}})}$ is created, spatially cropped to isolate the equipment region using the bounding boxes from $o_{t,i}$ associated with $e$. This reduces background clutter and emphasizes regions relevant to activity classification [21].

**Activity classification** The activity classification module processes video clips generated for each equipment instance. Each clip forms a spatiotemporal volume of frames that captures the equipment's motion during a candidate activity interval. The objective is to determine whether the clip contains a handling primitive. To this end, an activity classifier (AC) is applied to each clip and produces a predicted label $\hat{a} \in \{\text{grasp}, \text{release}, \text{none}\}$, along with a representative timestamp $t_{\text{act}} \in (t_{\text{start}}, t_{\text{end}})$:

$$\text{AC}(V^e_{(t_{\text{start}}, t_{\text{end}})}) = (\hat{a}, t_{\text{act}}). \tag{5}$$

During operation, the activity $a_t^e$ of equipment $e$ over the interval $[t_{\text{start}}, t_{\text{end}}]$ cannot be assigned until the clip is finalized and processed. Once the classification result becomes available, the predicted label and timestamp are used to update the activity sequence:

$$a_t^e = \begin{cases} \hat{a}, & \text{if } t = t_{\text{act}} \\ \text{none}, & \text{otherwise} \end{cases} \quad \text{for } t \in [t_{\text{start}}, t_{\text{end}}]. \tag{6}$$

The handling primitives are defined as *grasp*, *release*, and *none*. Grasp and release represent the elementary contact transitions that mark the initiation and termination of interaction between equipment and workpiece. None serves as a complementary class that indicates the absence of either transition, encompassing both periods of loaded transport and intervals without engagement.
The classifier is implemented on a 3D-CNN backbone that extracts spatiotemporal features from each input clip. To emphasize regions that are critical to interaction, such as crane grippers, we introduce a Keypoint-Guided Attention (KPGA) mechanism (Fig. 3). First, the spatiotemporal attention map $A_{ST}$ reweights the intermediate spatiotemporal features. Given an intermediate feature map $F_{\text{in}} \in \mathbb{R}^{T' \times C' \times H' \times W'}$, the attention maps are computed as

$$A_S = \sigma(\text{conv}(\text{pooling}(F_{\text{in}}))) \in \mathbb{R}^{T' \times H' \times W'}, \tag{7}$$

$$A_T = \sigma(\text{fc}(\text{pooling}(A_S)) \in \mathbb{R}^{T'}, \tag{8}$$

$$A_{ST} = A_S \otimes A_T \in \mathbb{R}^{T' \times H' \times W'}, \tag{9}$$

$$F_{\text{out}} = F_{\text{in}} \otimes A_{ST} \in \mathbb{R}^{T' \times C' \times H' \times W'}. \tag{10}$$

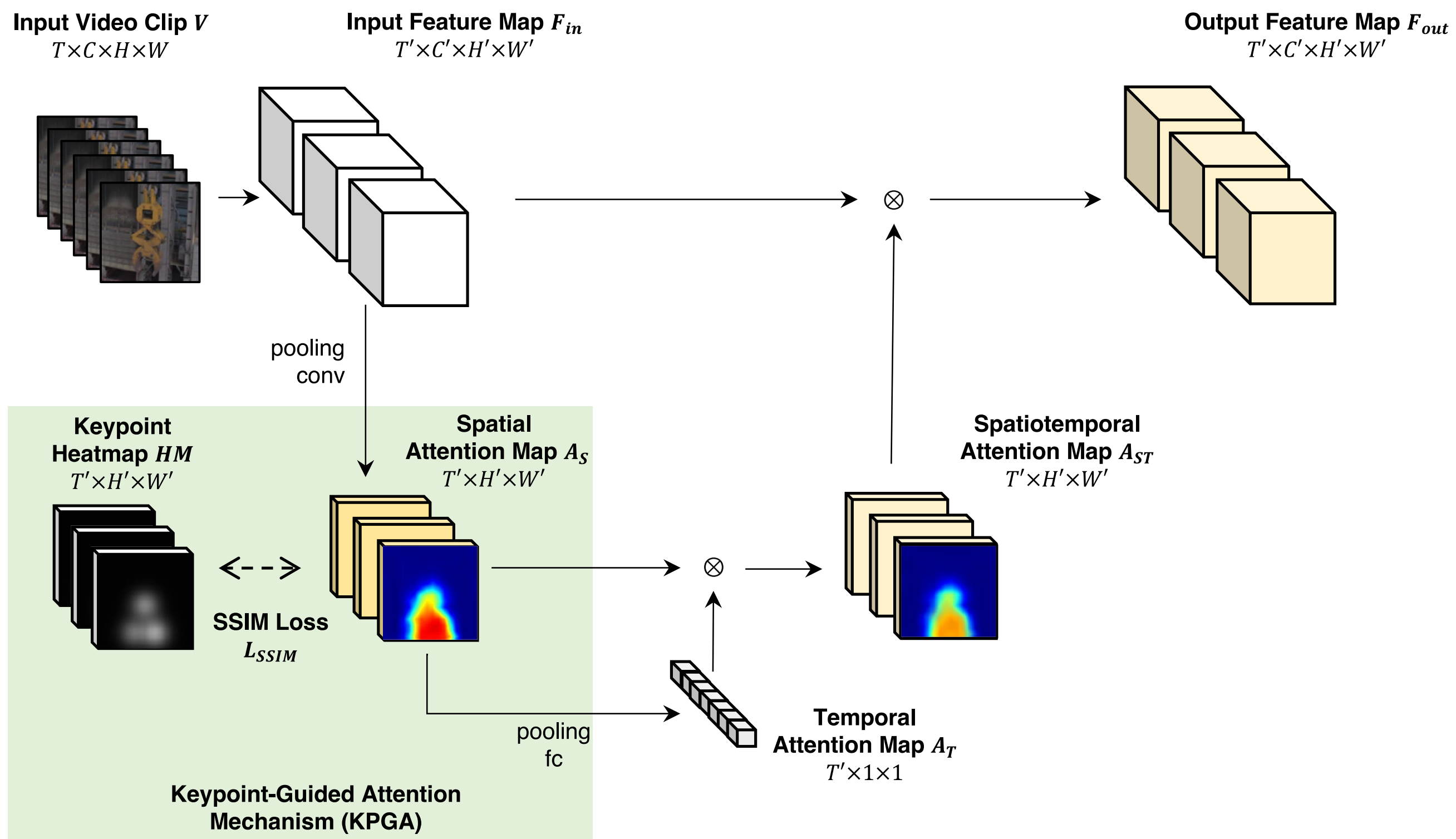

**Fig. 3** A Keypoint-Guided Attention (KPGA) mechanism for activity classification, which applies spatiotemporal attention guided by detected keypoints to enhance focus on regions relevant to crane–workpiece interactions

Then, to ensure spatial attention remains aligned with interaction cues, KPGA introduces a structural similarity loss $L_{\text{SSIM}}$ between the learned spatial attention map $A_S$ and the keypoint-derived heatmap $HM$. The reweighted feature map $F_{\text{out}}$ is passed through a classification head to produce a probability distribution $\text{prob}_{\text{out}} \in \mathbb{R}^3$ over the primitive labels. The overall training objective is expressed as

$$L_{\text{Total}} = L_{\text{CE}} + \lambda \cdot L_{\text{SSIM}}, \tag{11}$$

where $L_{\text{CE}}$ denotes the cross entropy loss, $\lambda$ balances classification accuracy with attention alignment, and $L_{\text{SSIM}}$ follows [22]. The representative timestamp $t_{\text{act}}$ is further refined using kernel density estimation over the equipment trajectory, with the peak density indicating the most likely instant of the primitive.

**Workpiece localization** Workpiece localization constitutes the interpretation stage of the framework, where perception outputs are transformed into semantically meaningful states of equipment and workpieces. This stage is realized in an event-driven manner through coupled finite state machines (FSMs). Events derived from estimated equipment activities are validated by spatial and contextual guards, and they drive transitions in the equipment FSMs while being propagated to the associated workpiece FSMs. As a result, both semantic states and physical coordinates of workpieces are updated continuously. In this formulation, event detection itself serves as the localization mechanism: each validated event directly determines the handling state and location of a workpiece.

Events provide an interpretable abstraction over raw perception and are derived from detected crane activity primitives. A `Grasp` event is generated when a workpiece lies within a tolerance distance of the crane's coordinate at the same time that a grasp primitive is predicted. A `Release` event occurs when a release primitive is predicted, signifying deposition of the carried workpiece at the crane's current coordinates. A `None` event indicates that no change in state occurs. Additional events ensure robustness and lifecycle management. `Fail` and `Recover` are triggered when inconsistent activity sequences are observed. For example, if a `Release` is detected without a preceding `Grasp`, the crane and the workpiece temporarily fall into an invalid configuration; recovery then restores them to their last consistent situation. `Incoming` and `Outgoing` events manage initialization and termination, marking the entry of a workpiece into the system and the completion of its lifecycle when the operation is done. These events are shared by both equipment and workpieces, ensuring synchronized evolution.

Each equipment FSM evolves over the states {Idle, Holding, Error}, while each workpiece FSM evolves over the states {Placed, Carried}. A coupling map links each equipment instance to the workpiece it is currently handling, thereby supporting coordinated transitions between the two layers. Localization is realized through state-conditioned coordinate updates: when a workpiece is in the Carried state, its coordinates follow that of the handling equipment, whereas in the Placed state, the coordinates remain fixed at its last valid coordinate. Lifecycle events guarantee consistent initialization and termination, yielding coherent workpiece trajectories throughout the yard.

The unified FSM specification, summarized in Fig. 4 and Table 1, constitutes the interpretable core of the proposed framework. Its detailed algorithmic realization is provided in Algorithm 1, which is presented in Appendix. At each timestep, equipment states are updated based on primitive activities, events are validated through spatial and contextual guards, and workpiece states and coordinates are revised accordingly.

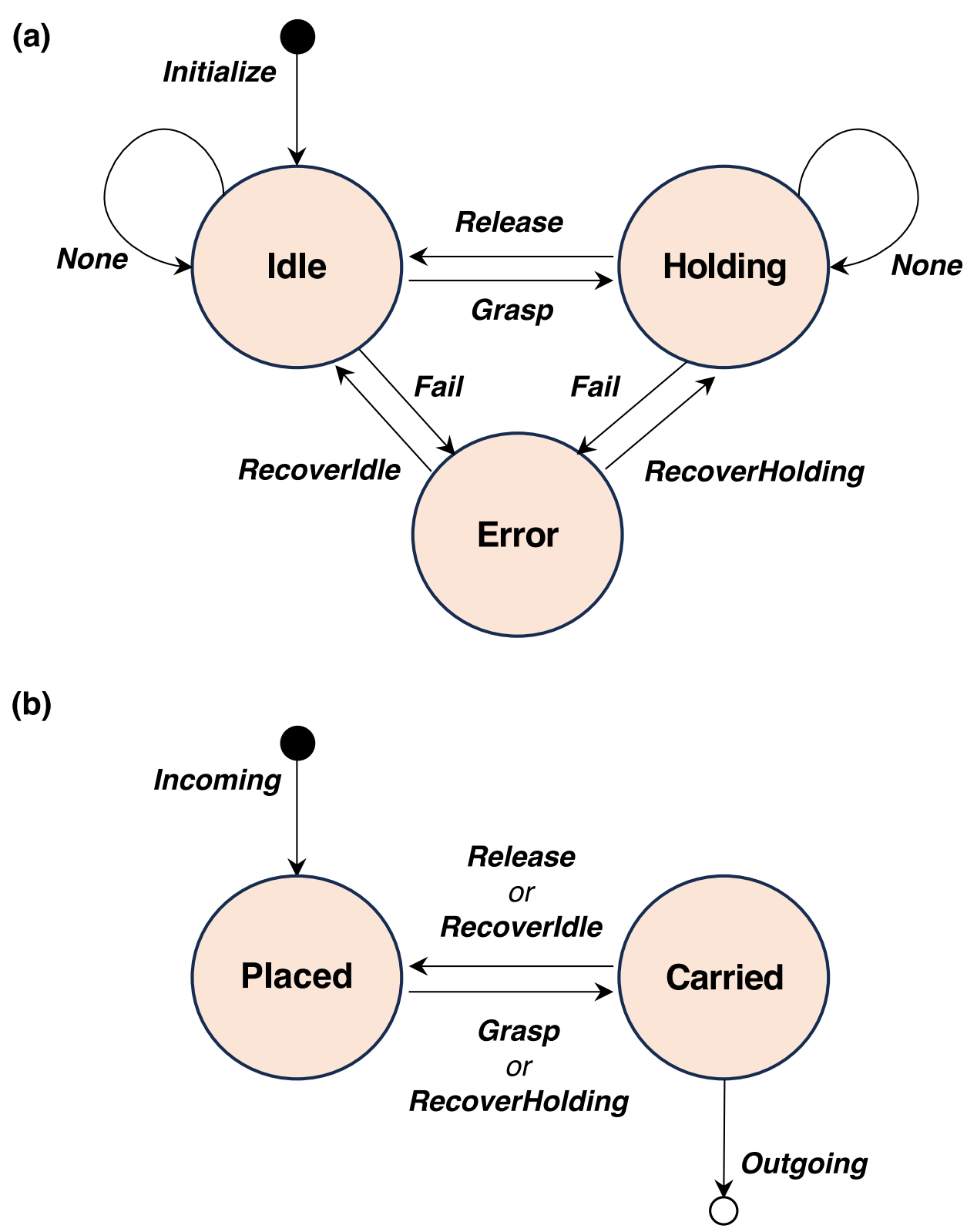

**Fig. 4** Event-driven finite state machines (FSMs) for (**a**) equipment and (**b**) workpieces

**Table 1** Core transition rules for unified FSMs

| Event | Equipment FSM | Workpiece FSM |
|---|---|---|
| Grasp | Idle → Holding | Placed → Carried |
| Release | Holding → Idle | Carried → Placed |
| None | Idle → Idle; Holding → Holding | Placed → Placed; Carried → Carried |
| Incoming | N/A | Initialized as Placed |
| Outgoing | N/A | Carried → Terminated |
| Fail | Idle/Holding → Error | No change |
| RecoverIdle | Error → Idle | Carried → Placed |
| RecoverHolding | Error → Holding | Placed → Carried |

notes: `Fail` suspends equipment–workpiece coupling. `RecoverIdle` and `RecoverHolding` re-synchronize the workpiece at its last valid state

## 4 Experimental validation

This section reports experimental validations conducted to evaluate the performance and feasibility of the proposed framework, which integrates two core stages: Equipment Tracking and Workpiece Localization. To reflect their distinct objectives, validations were organized into two modes: offline validation for model development and accuracy assessment, and online validation for end-to-end evaluation under operational conditions.

### 4.1 Setup

Validations were carried out at a hot forging facility comprising a yard, five furnaces, and a forging station serviced by two overhead cranes. The yard, which defines the global coordinate system for localization, measures approximately 67 m in width, 22.8 m in length, and 18.6 m in height. Each furnace (5.8 m × 11.1 m × 7.25 m) is positioned along the yard boundary and temporarily shares the global coordinate system when its door is opened for transfers. Workpieces enter the yard as raw materials, are heated in furnaces, forged, and cooled before possibly repeating the cycle; all inter-station transfers are performed by the two overhead cranes, which are restricted to carrying a single workpiece at a time for safety. This workflow exemplifies the typical workpiece flow in hot forging operations and provided the context for evaluating the proposed localization framework (Fig. 5 (a)).

A multi-camera system was installed at elevated positions to monitor crane activities and workpiece transfers. Two longitudinal cameras covered the yard and furnace entrances along the primary axis, while one transverse camera monitored across the furnace, ensuring complete visual coverage. All cameras operated at 1 FPS, relying on existing

(a)

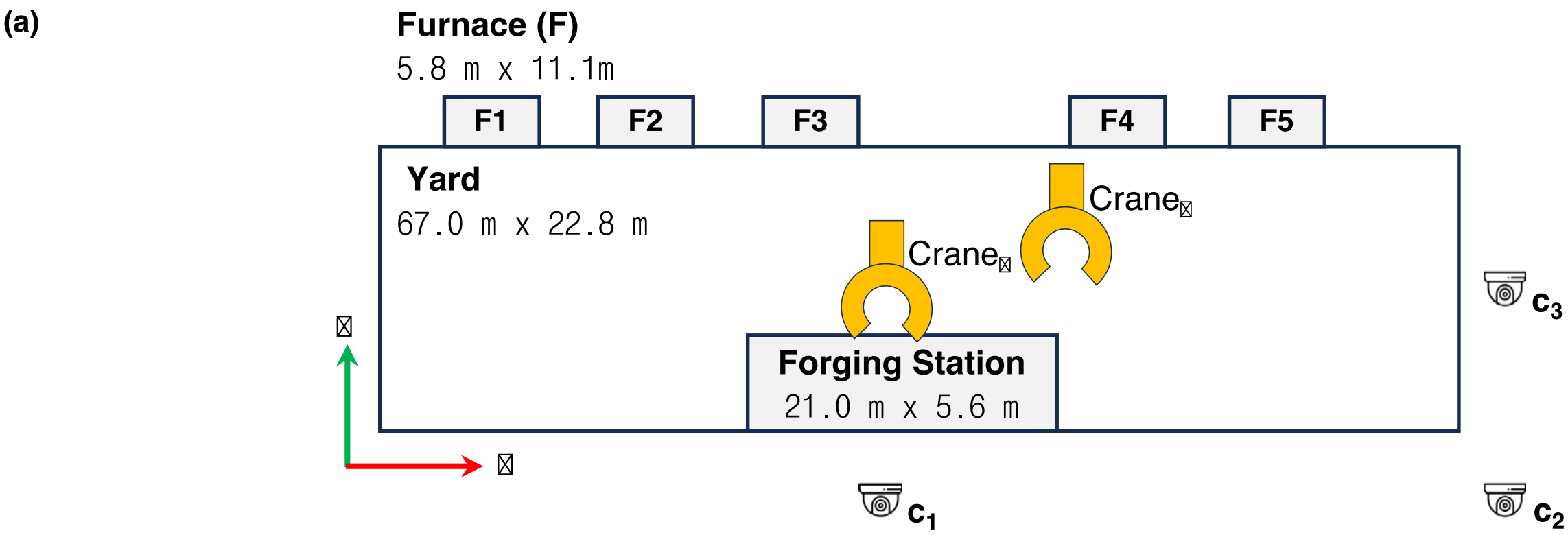

(b)

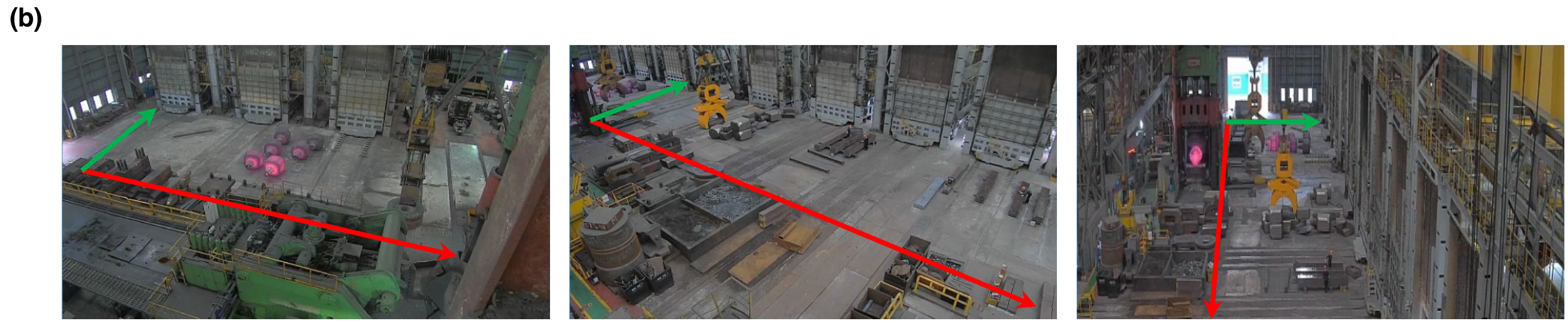

**Fig. 5** Layout of CCTV cameras ($c_1$, $c_2$, $c_3$) and their coverage of the factory, showing (**a**) camera placements on the 2D floorplan and (**b**) corresponding fields of view

factory lighting without additional calibration (Fig. 5 (b)). Homography matrices and height-displacement vectors were derived once from floorplan measurements and the camera configurations.

Offline validation employed image and video datasets collected between July and September 2024, and was used to establish baseline performance of the perception modules. Online validation was conducted across three representative deployment periods: October 1–4 (Period 1), October 15–18 (Period 2), and October 28–31 (Period 3), 2024, thereby evaluating the complete event-driven localization framework during live operations.

All experiments were executed on a workstation equipped with an Intel Xeon Silver CPU, 32 GB RAM, and an NVIDIA A5000 GPU. For consistency across modules, the displacement threshold $\theta$ and the spatial proximity threshold $\delta$ were both set to 800 mm, corresponding to the minimum workpiece length.

## 4.2 Offline validation

### 4.2.1 Pose estimator

The pose estimator was implemented using YOLO-Pose [23], which detects both bounding boxes and keypoints in a single inference step. The model was trained on a custom dataset collected from the operational environment to adapt to the domain-specific appearance and spatial configuration of the equipment. For temporal consistency across frames, detections were associated using the Simple Online Realtime Tracker (SORT) algorithm [24].

A total of 1,851 images were collected from randomly sampled video frames obtained from three camera viewpoints between July and September 2024. To accelerate dataset preparation, a hybrid manual–pseudo labeling procedure was adopted: 688 images were manually annotated and used to pretrain a YOLO-Pose model, while the remaining 1,168 images were first annotated by the pretrained model and subsequently revised by human annotators. Dataset statistics across cameras and equipment classes are summarized in Table 2. Training details and hyperparameters are provided in the Appendix.

Annotations included bounding boxes and class-specific keypoints. Keypoints were defined based on each equipment's role in workpiece handling and geometric characteristics. For cranes (Fig. 6 (a)), three keypoints were designated: the gripper center, the left arm tip, and the right arm tip. The representative point for each crane was computed as the midpoint between the two arm tips, approximating the actual grasping region and serving as a positional reference for downstream tasks. For furnace doors (Fig. 6 (b)), two keypoints were defined at the left and right door edges, with the midpoint providing a representative position. This definition was applied consistently across the dataset to ensure reliable localization and event detection.

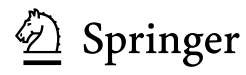

**Table 2** Training and validation image counts per camera and equipment class

| Camera | Class | Train Images | Val Images |
|---|---|---|---|
| $c_1$ | Total | 1023 | 71 |
| | Crane | 987 | 60 |
| | Furnace Door | 3055 | 215 |
| $c_2$ | Total | 735 | 60 |
| | Crane | 1012 | 66 |
| | Furnace Door | 4717 | 393 |
| $c_3$ | Total | 264 | 25 |
| | Crane | 387 | 40 |
| | Furnace Door | – | – |

*Notes:* furnace doors are not present in the $c_3$ viewpoint. Instance counts exceed image counts as multiple objects may appear per image

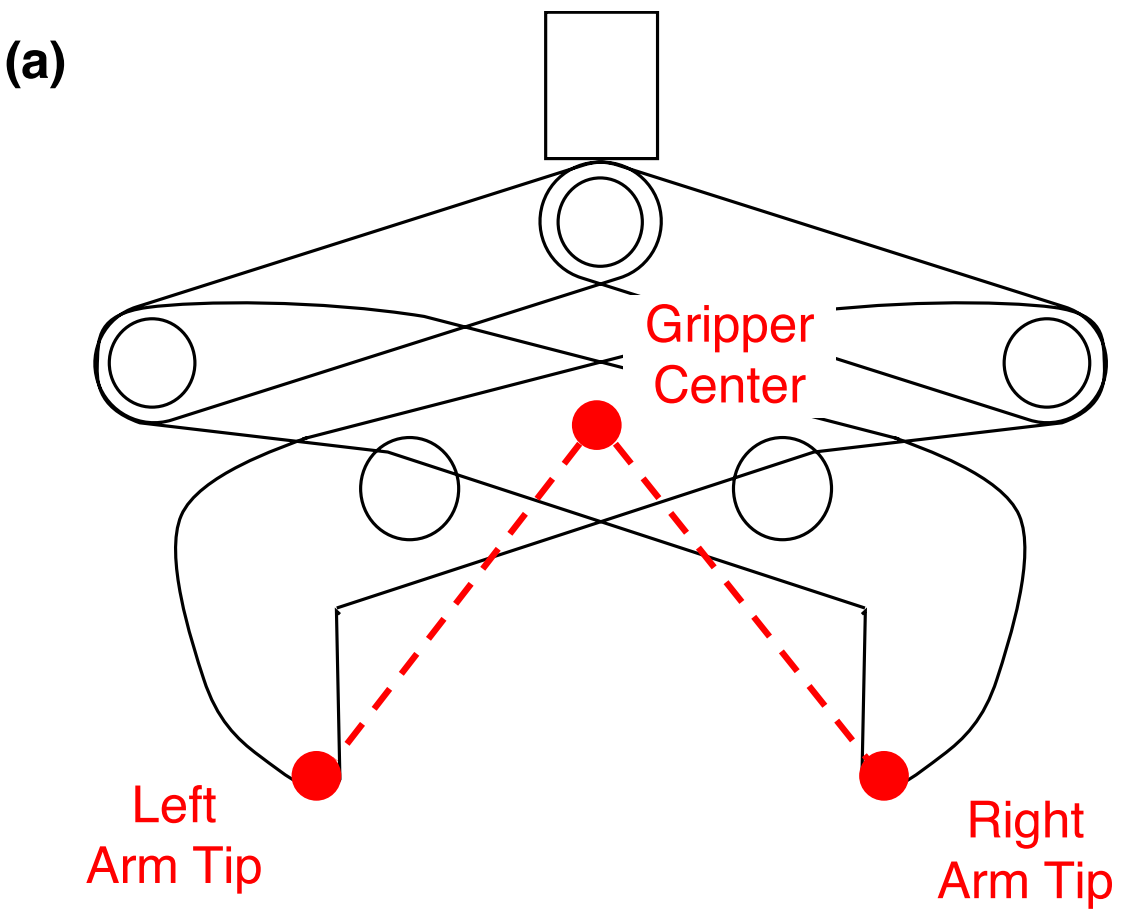

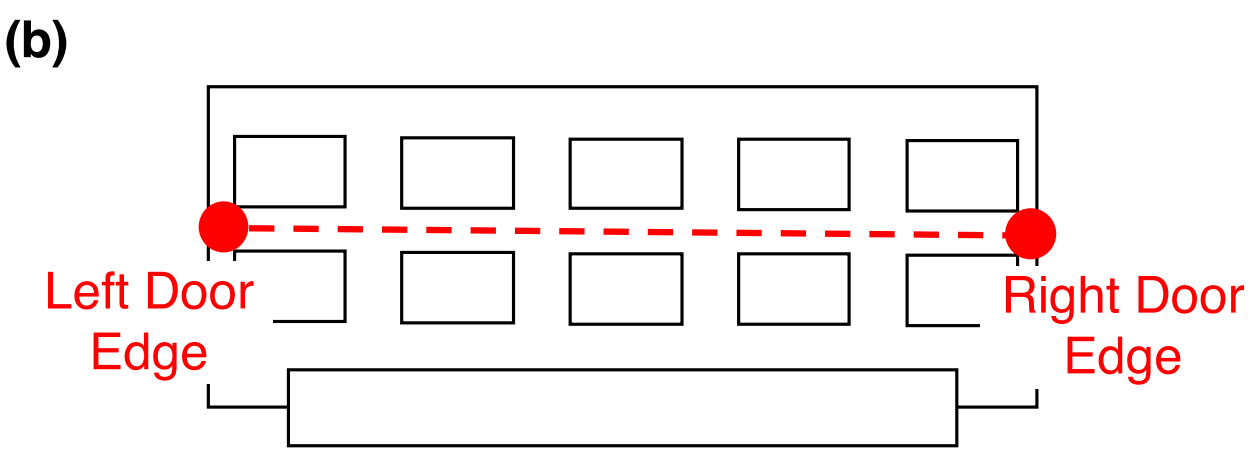

**Fig. 6** Keypoint annotation scheme for (**a**) crane and (**b**) furnace door

Validation results are summarized in Table 3. Across both equipment classes, the estimator achieved near-perfect detection and pose accuracy, with mAP@0.5 values exceeding 0.98 for both bounding boxes and keypoints. Performance was consistent across cameras, and inference speed averaged 49.5 FPS on the target GPU, confirming real-time capability for multi-camera deployments.

These results demonstrate that the estimator provides sufficiently robust and accurate equipment localization to support the subsequent activity recognition and workpiece localization stages.

**Table 3** Validation performance of bounding box and keypoint detection by equipment class

| Class | Metric | Box Metric | Pose Metric |
|---|---|---|---|
| Crane | Precision | 0.973 | 0.988 |
| | Recall | 0.973 | 0.988 |
| | mAP@0.5 | 0.988 | 0.988 |
| Furnace Door | Precision | 1.000 | 0.995 |
| | Recall | 1.000 | 0.995 |
| | mAP@0.5 | 0.995 | 0.995 |

### 4.2.2 Activity classifier

The activity classifier was evaluated to determine its ability to recognize the handling primitives (*grasp*, *release*, *none*) prior to deployment. A total of 4,379 video clips were collected and annotated between July and September 2024 using the activity proposal mechanism described in Section 3.2.2, capturing realistic operational scenarios. Dataset composition by activity class and data split is summarized in Table 4. Training procedures and architectural specifications of the evaluated backbones are provided in the Appendix.

Three representative 3D-CNN backbones—R(2+1)D [25], R3D [26], and X3D [27]—were examined under plain, NST-only, and KPGA-enhanced configurations, with the overall architecture shown in Fig. 7. Two transformer-based models (MViTv2 [28] and Swin3D [29]) were also included as external baselines.

Validation results are presented in Table 5. Across all backbones, KPGA consistently improved performance, with the R(2+1)D+KPGA model achieving the highest accuracy at 94.2%. The transformer baselines performed competitively but did not surpass the KPGA-enhanced 3D-CNNs.

Beyond accuracy, KPGA provided qualitative benefits by generating attention maps that revealed where and when the model focused during activity recognition. As illustrated in Fig. 8, KPGA guided attention toward the crane gripper and its interaction with the workpiece, while suppressing irrelevant background motion. The temporal evolution of these maps aligned with activity phases, indicating whether the

**Table 4** Number and average length of collected video clips by activity class and data split

| Class | Split | Number of Clips | Average Clip Length (s) |
|---|---|---|---|
| Grasp | Train | 989 | 59.15 |
| | Val | 243 | 58.69 |
| Release | Train | 928 | 41.50 |
| | Val | 239 | 41.87 |
| None | Train | 963 | 37.38 |
| | Val | 241 | 38.04 |

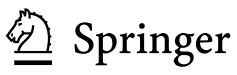

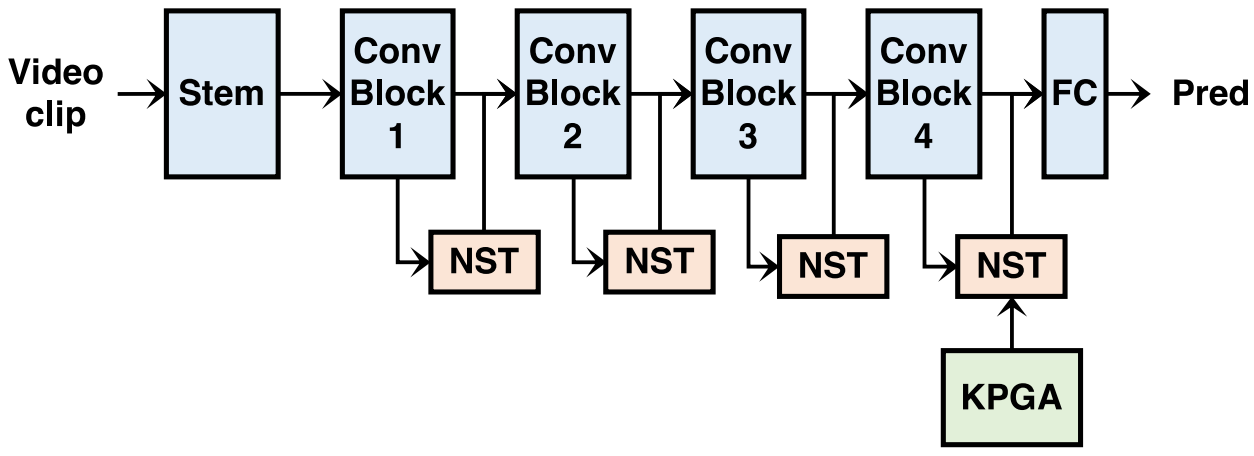

**Fig. 7** Architecture of the activity classifier integrating a ResNet-34 [30] backbone, NST modules, and the KPGA mechanism

**Table 5** Validation accuracy (%) of activity classifiers across 3D-CNN backbones and SSIM loss coefficients, with comparisons between plain, NST-only, and KPGA-enhanced configurations

| Backbone | Plain | NST-only | KPGA | | |
|---|---|---|---|---|---|
| | | | $\lambda = 0.1$ | $\lambda = 1$ | $\lambda = 5$ |
| R(2+1)D | 90.57 | 89.50 | 91.16 | **<u>94.20</u>** | 93.10 |
| R3D | 91.27 | 92.37 | 91.07 | 92.27 | **93.10** |
| X3D | 91.97 | 91.17 | 90.57 | **93.97** | 91.07 |
| MViTv2 | **92.27** | - | - | - | - |
| Swin3D | **89.77** | - | - | - | - |

notes: bold values indicate the best accuracy for each backbone and the globally best accuracy across all settings is additionally underlined

crane was actively holding or releasing a workpiece. These visualizations not only improve confidence in the classifier's decisions but also demonstrate that KPGA enhances interpretability—a crucial property for event-driven systems deployed in industrial contexts.

Taken together, the results confirm that KPGA strengthens both classification accuracy and interpretability, providing a reliable foundation for event-driven localization in live operations.

### 4.3 Online validation

Online validation was conducted to evaluate the complete Workpiece Localization stage under real operating conditions. A total of 23 workpieces were tracked from `Incoming` registration to `Outgoing` dispatch across three deployment periods, with two overhead cranes performing handling operations. This setting allowed assessment of event detection, state consistency, localization accuracy, and the operational insights enabled by the framework.

During implementation, the equipment FSM occasionally entered the Error state due to the detection of invalid events. When this occurred, the system required a revision process to restore propagation of crane and workpiece states. In our validation, corrections were applied manually.

**Event detection results** Across all periods, a total of 160 `Grasp`, 158 `Release`, and 195 `None` events were detected. In total, 34 `Fail` events were recorded—32 from $Crane_1$ and 2 from $Crane_2$. Each was subsequently corrected, producing 19 `RecoverIdle` and 15 `RecoverHolding` events, which restored the FSMs to valid states. In parallel, 23 workpieces were tracked to completion, confirming consistent lifecycle management.

**Event detection accuracy (EDA)** EDA quantifies the temporal and semantic correctness of detected events relative to manually annotated ground truth. Each annotation specified the event type, timestamp, and associated entities (e.g., crane or workpiece ID). For consistency, `RecoverIdle` and `RecoverHolding` were treated as equivalent to `Release` and `Grasp`, respectively. Let $\mathcal{E}_{\mathrm{dt}}$ and $\mathcal{E}_{\text{gt}}$ denote the sets of detected and ground truth events. A detected event $\varepsilon_{\text{dt}} \in \mathcal{E}_{\text{dt}}$ with timestamp $t_{\text{dt}}$ was matched to a ground truth event $\varepsilon_{\text{gt}} \in \mathcal{E}_{\text{gt}}$ with timestamp $t_{\text{gt}}$ if (i) their types and entities coincided, and (ii) $|t_{\text{dt}} - t_{\text{gt}}| \leq \Delta t$. Once matched, both events were removed to enforce one-to-one correspondence. EDA was then defined as

$$\text{EDA}(\Delta t) = \frac{\#\ \text{matched pairs within } \Delta t}{|\mathcal{E}_{\text{gt}}|}. \tag{12}$$

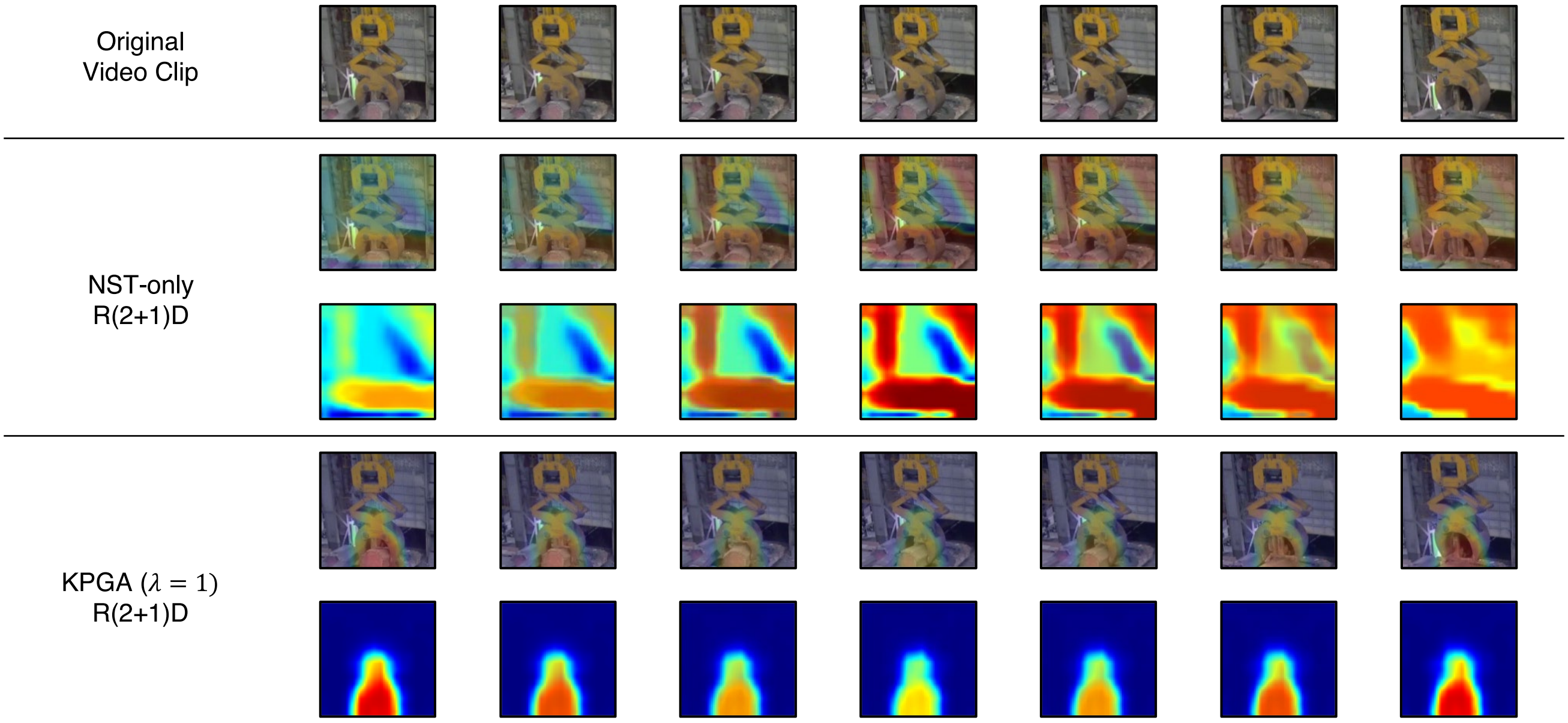

**Fig. 8** Comparison of attention maps between NST baseline and KPGA-enchanced R(2+1)D

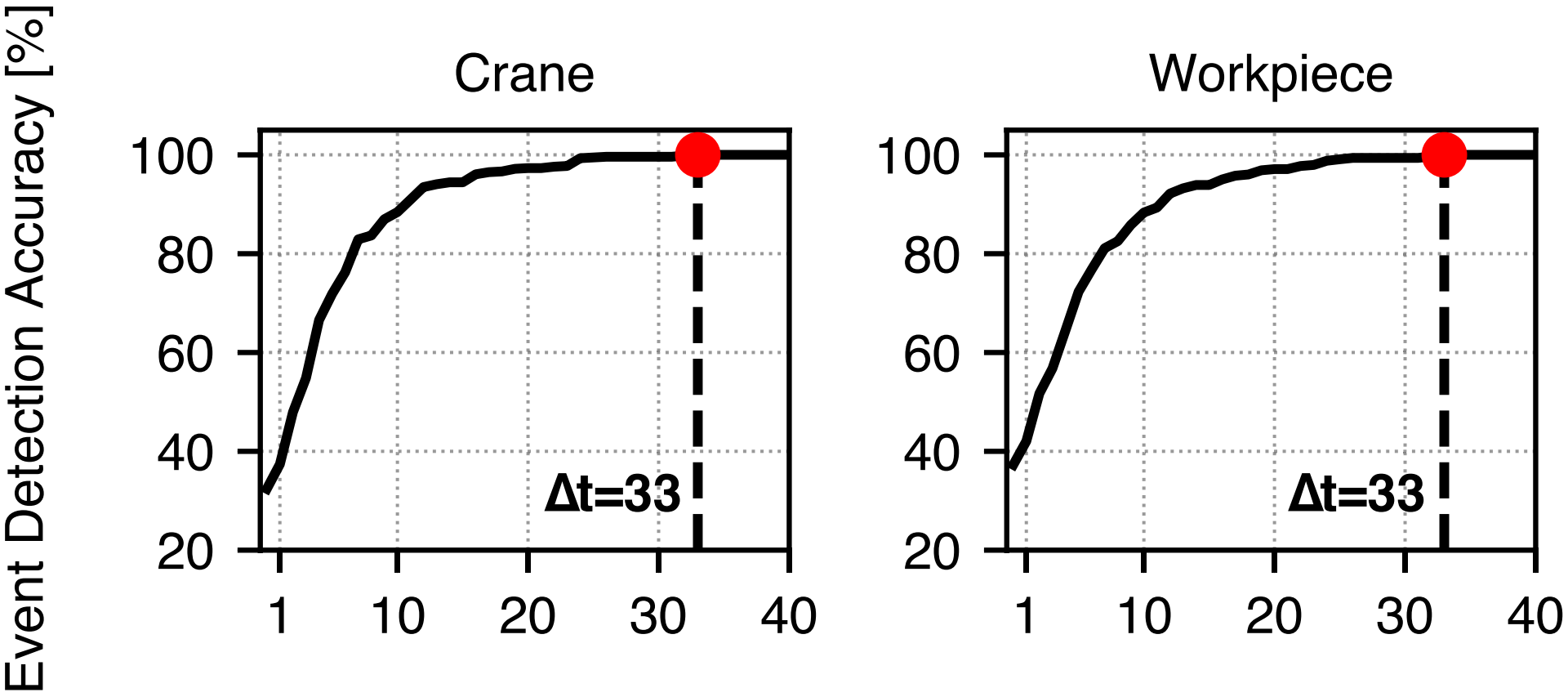

**Fig. 9** Event Detection Accuracy (EDA) curves for crane and workpiece over time tolerances

This metric expresses the proportion of ground truth events correctly detected within the specified temporal tolerance. As shown in Fig. 9, EDA reached 100% at $\Delta t = 33$ seconds, demonstrating precise temporal and semantic alignment.

**Event detection latency** End-to-end responsiveness is governed by a four-step *critical path* that begins once an activity interval closes; upstream steps (frame capture/decoding, pose estimation, tracking/association, and floor localization) run continuously at 1 FPS and overlap with acquisition, so they do not add wall-clock delay. Mean latencies of each step are summarized in Table 6. The Activity Proposal step dominates (16.9 seconds). Clip Generation contributes $\approx 0.5$ seconds, while the FSM Update and Activity Classification are negligible ($< 0.1$ seconds). The measured mean end-to-end latency is 17.4 seconds, comfortably below the minimum observed 35 seconds between consecutive handling events, thus achieving near real-time performance.

**Table 6** Mean latencies of the four-step critical path in the event-detection pipeline

| Stage | Mean latency (sec) |
|---|---|
| Activity Proposal | 16.9 |
| Clip Generation | 0.5 |
| Activity Classification | $< 0.1$ |
| FSM Update | $< 0.1$ |
| *End-to-end (measured)* | 17.4 |

**Localization error** Localization accuracy was assessed at `Release` events where ground truth coordinates were available. Across 167 events, the mean error was 317.8 mm with a standard deviation of 236.2 mm. In 97.6% of cases, the error remained below 800 mm, corresponding to the minimum workpiece length (Fig. 10). Outliers were primarily caused by keypoint detection inaccuracies during online inference but were rare and did not affect overall consistency.

Additional details of online validation, including the correction procedure, per-crane and per-workpiece event statistics, and extended EDA curves, are provided in the Appendix.

**Visualization and operational insights** Beyond validation metrics, the framework generated structured outputs for process-level analysis. Figure 11 shows a representative workpiece trajectory, highlighting transfer routes and staging behavior. FSM-derived state histories supported the construction of Gantt charts of workpiece residence times (Fig. 12), revealing practices such as the last workpiece of a batch bypassing the yard. Equipment utilization analysis further showed workload imbalance, with $Crane_1$ executing 300 handling operations (2.75 hours active) compared to 34 by $Crane_2$ (0.31 hours). Such imbalances highlight opportunities for workload balancing, throughput improvement, and preventive maintenance planning.

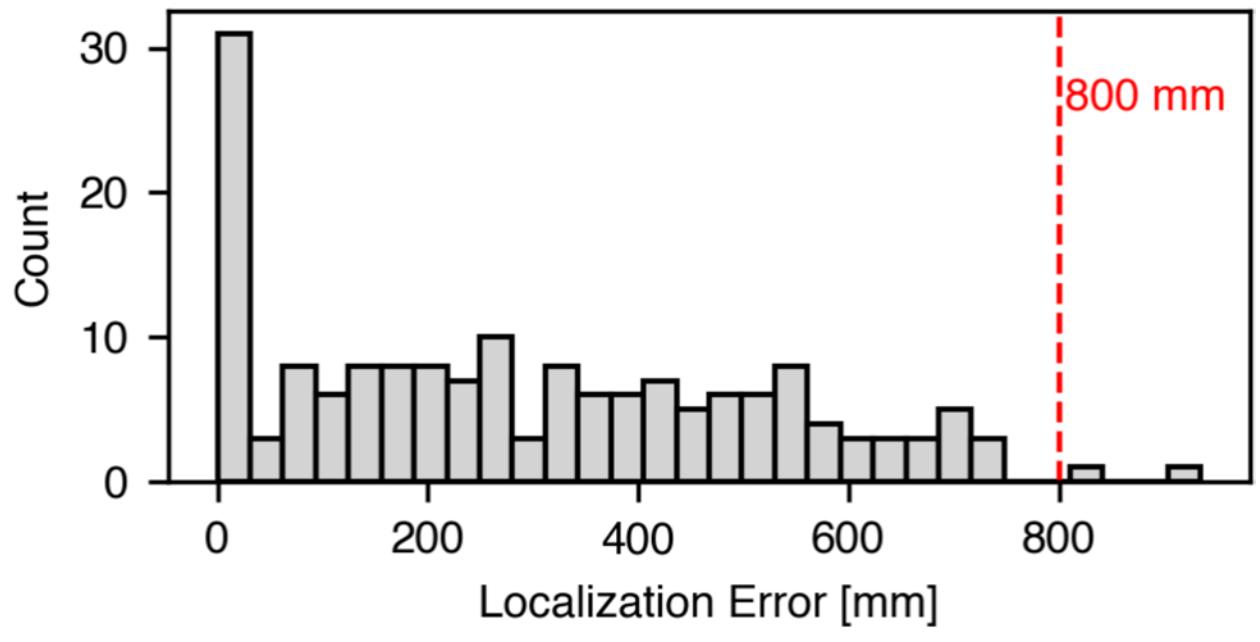

**Fig. 10** Distribution of localization errors measured at `Release` events

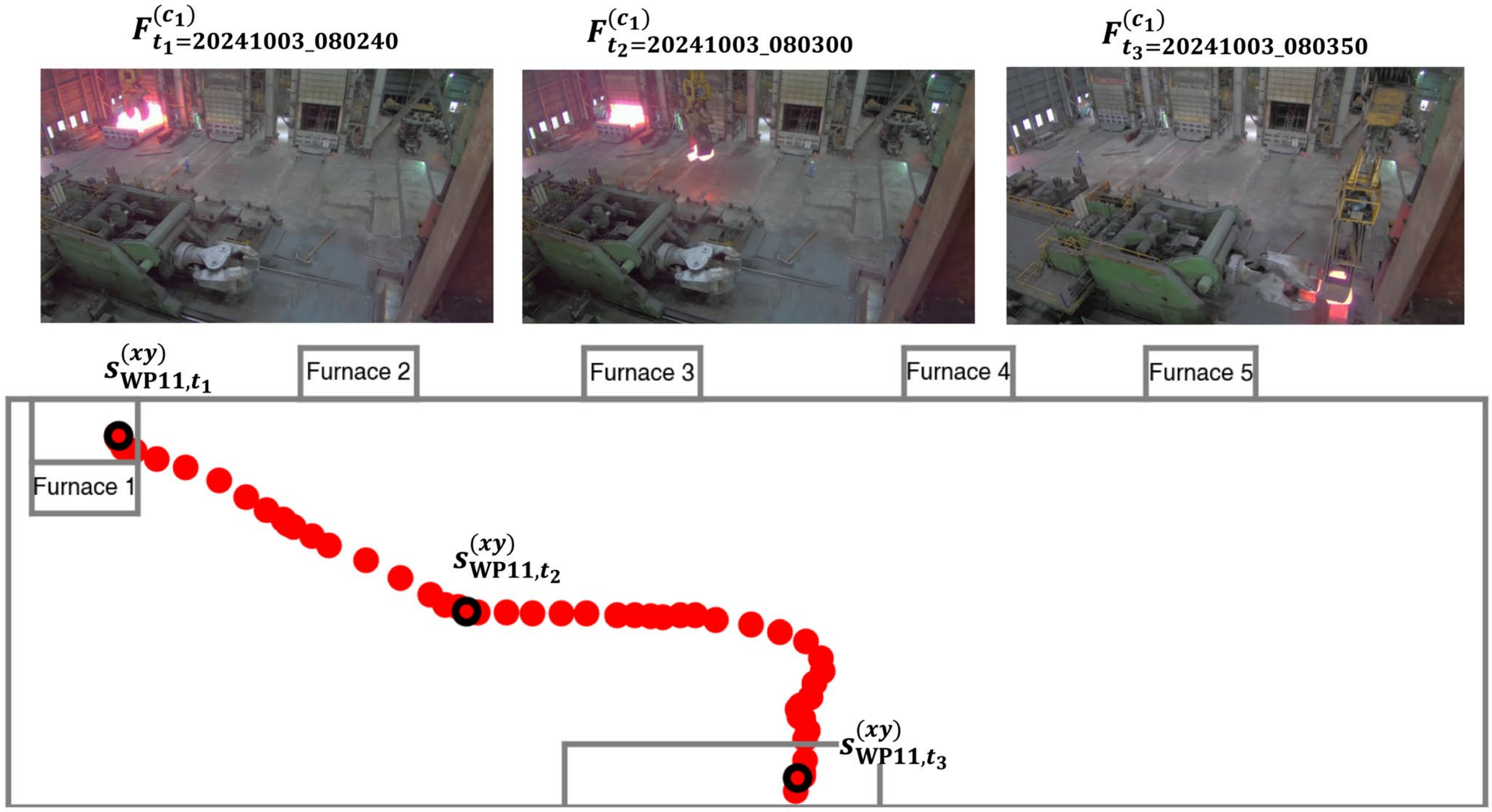

**Fig. 11** Transfer of workpiece WP11 from Furnace 1 to the Forging station, showing camera $c_1$ views at three time points (top) and corresponding floorplan-space coordinates $s^{(xy)}_{\mathrm{WP11},t}$ over time (bottom)

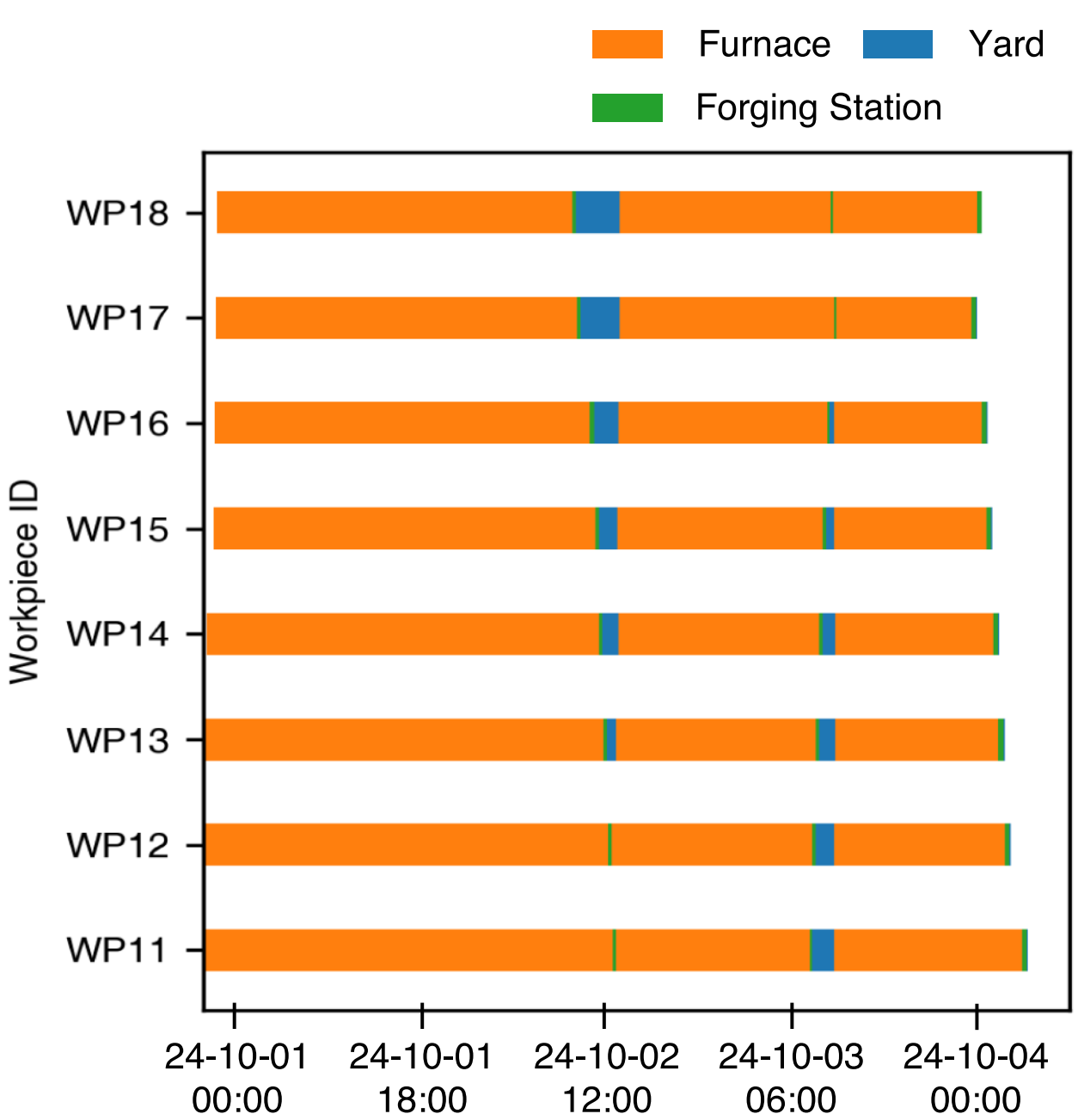

**Fig. 12** Gantt chart of FSM-derived workpiece residence times at the Yard, Furnace, and Forging Station

Collectively, these results demonstrate that the proposed framework extends beyond accurate continuous localization to serve as a comprehensive tool for operational intelligence. By transforming spatiotemporal data into interpretable visualizations and structured analytics, the system enables effective monitoring of workpiece flow, early detection of inefficiencies, quantitative evaluation of equipment usage, and identification of improvement opportunities.

## 5 Conclusion

This study presented an equipment-centric workpiece localization framework for hot forging environments that operates in near real time. By continuously estimating floorplan-space 3D coordinates of handling equipment and detecting workpiece-handling activities from multi-view video streams, the system inferred workpiece coordinates indirectly through an event-driven FSM structure. The proposed framework integrated three key components: multi-view equipment tracking via keypoint-based pose estimation and spatial projection, activity classification enhanced by the Keypoint-Guided Attention (KPGA) mechanism, and synchronized equipment and workpiece FSMs for logical state updates. Validation in an operational factory demonstrated high event detection accuracy within a 33-second tolerance window, a mean localization error of 317.8 mm, and a system latency suitable for near real-time operation. In addition to localization, the structured outputs enabled visualization of workpiece transfers, residence time analysis across the Yard, Furnaces, and Forging Station, and evaluation of crane usage. These capabilities provided operational insights such as material congestion near specific stations, batch-based transfer behaviors, and crane workload imbalance.

While the framework demonstrated robust performance in practice, several areas remain for improvement. The current activity proposal module, based on low-displacement intervals, is effective but can introduce latency and occasionally miss subtle transitions; more adaptive buffering or streaming-based recognition could reduce this overhead and improve responsiveness. During validation, FSM revision was performed manually to resolve invalid transitions caused by misclassified activities; automating this recovery through anomaly-aware detection or rule-based correction would reduce reliance on human intervention. The structured outputs also present opportunities for integration with higher-level manufacturing systems, where linking inferred workpiece states with a Manufacturing Execution System (MES) could enable enhanced traceability and scheduling optimization.

Beyond the specific case study, the proposed framework represents a shift from traditional workpiece-centric approaches to an equipment-centric paradigm for continuous localization. The unified FSM specification extends naturally to multiple concurrent equipment units—such as additional cranes or robotic manipulators—by maintaining coupled state machines for each handler–workpiece pair. As the number of handlers increases, coordination complexity grows, but the shared event-driven logic ensures consistency and interpretability. Misclassification recovery can be supported by cross-checks among equipment FSMs, and KPGA-driven activity recognition can adapt to different equipment types given appropriate keypoint definitions. These characteristics suggest that the proposed approach can be applied not only to larger hot forging shops but also to other industrial domains involving frequent material transfers and concurrent handling, contributing to more intelligent, data-driven, and traceable manufacturing operations.

## Appendix

### Localization algorithm

The described FSM logic is implemented through localization algorithm (Algorithm 1). At each timestep, the system integrates the equipment coordinate $s_{t,e}$, the predicted activity $a_{t,e}$, and spatial proximity measures to detect events and trigger FSM transitions accordingly.

**Algorithm 1** Real-Time Workpiece Localization via FSM Revisions

1: **Input:** Equipment set $E'$, spatial proximity threshold $\delta$
2: **Initialize:**
3: $\text{FSM}^{e}_{\text{equipment}} \leftarrow S_{\text{Idle}}$ **for all** $e \in E'$
4: $W \leftarrow \emptyset$ ▷ Workpiece set initially empty
5: $H \leftarrow \{\}$ ▷ Equipment-to-workpiece map: $H[e] = w$ if $e$ is holding $w$
6: **for** each timestep $t$ **do**
7: **for** each newly registered workpiece $w$ **do**
8: $W \leftarrow W \cup \{w\}$; $\text{FSM}^{w}_{\text{workpiece}} \leftarrow S_{\text{Placed}}$
9: **end for**
10: **for** each equipment $e \in E'$ **do**
11: Observe floorplan-space coordinate $s_{t,e}$ and activity $a^{e}_{t}$
12: **if** $a^{e}_{t} = grasp$ **and** $\text{FSM}^{e}_{\text{equipment}} = S_{\text{Idle}}$ **then** ▷ Detect `Grasp` event
13: Identify nearest $w^{*} \in W$ within $\delta$ of $s_{t,e}$
14: $\text{FSM}^{e}_{\text{equipment}} \leftarrow S_{\text{Holding}}$; $\text{FSM}^{w^{*}}_{\text{workpiece}} \leftarrow S_{\text{Carried}}$; $H[e] \leftarrow w^{*}$
15: **else if** $a^{e}_{t} = release$ **and** $\text{FSM}^{e}_{\text{equipment}} = S_{\text{Holding}}$ **then** ▷ Detect `Release` event
16: $\text{FSM}^{e}_{\text{equipment}} \leftarrow S_{\text{Idle}}$; $\text{FSM}^{H[e]}_{\text{workpiece}} \leftarrow S_{\text{Placed}}$; $H[e] \leftarrow \text{None}$
17: **else if** $a^{e}_{t} = none$ **then** ▷ Detect `None` event
18: **continue**
19: **else** ▷ Detect `Fail` event
20: $\text{FSM}^{e}_{\text{equipment}} \leftarrow S_{\text{Error}}$
21: **end if**
22: **end for**
23: **for** each $w \in W$ **do**
24: **if** $\text{FSM}^{w}_{\text{workpiece}} = S_{\text{Carried}}$ **then**
25: Find $e$ such that $H[e] = w$ ; $s_{t,w} \leftarrow s_{t,e}$ ▷ Workpiece coordinate follows equipment
26: **else**
27: $s_{t,w} \leftarrow s_{t-1,w}$
28: **end if**
29: **if** workpiece $w$ crosses boundary **then**
30: Terminate $\text{FSM}^{w}_{\text{workpiece}}$
31: **end if**
32: **end for**
33: **end for**

### Implementation details

**Pose estimator training details** The YOLO-Pose model with a YOLOv11s backbone [31] was trained using input images resized to 640 × 480 pixels. Training employed the SGD optimizer with an initial learning rate of 0.01, linearly decayed to 0.0001, a batch size of 16, and 200 epochs. Data augmentation included random HSV adjustment, horizontal flipping, and mosaic augmentation to enhance robustness.

**Activity classifier training details** 3D-CNN backbones were trained under uniform conditions using the SGD optimizer with an initial learning rate of 0.01. All models

**Table 7** Architectural specifications of activity classifiers with R(2+1)D, R3D, and X3D backbones

| Backbone | R(2+1)D | R3D | X3D |
|---|---|---|---|
| Video Clip Shape $T \times C \times H \times W$ | $20 \times 3 \times 256 \times 256$ | $20 \times 3 \times 256 \times 256$ | $20 \times 3 \times 256 \times 256$ |
| Stem | $3 \times 7 \times 7$ conv, 64<br>$2 \times 2$ Max Pooling | $3 \times 7 \times 7$ conv, 64<br>$2 \times 2$ Max Pooling | $1 \times 3 \times 3$ conv, 24 |
| Conv Block 1 | $\begin{bmatrix} 1 \times 3 \times 3 \text{ conv, } 64 \\ 3 \times 1 \times 1 \text{ conv, } 64 \end{bmatrix} \times 3$ | $\begin{bmatrix} 3 \times 3 \times 3 \text{ conv, } 64 \\ 3 \times 3 \times 3 \text{ conv, } 64 \end{bmatrix} \times 3$ | $\begin{bmatrix} 1 \times 1 \times 1 \text{ conv, } 54 \\ 3 \times 3 \times 3 \text{ conv, } 54 \\ 1 \times 1 \times 1 \text{ conv, } 54 \end{bmatrix} \times 3$ |
| Conv Block 2 | $\begin{bmatrix} 1 \times 3 \times 3 \text{ conv, } 128 \\ 3 \times 1 \times 1 \text{ conv, } 128 \end{bmatrix} \times 4$ | $\begin{bmatrix} 3 \times 3 \times 3 \text{ conv, } 128 \\ 3 \times 3 \times 3 \text{ conv, } 128 \end{bmatrix} \times 4$ | $\begin{bmatrix} 1 \times 1 \times 1 \text{ conv, } 108 \\ 3 \times 3 \times 3 \text{ conv, } 108 \\ 1 \times 1 \times 1 \text{ conv, } 108 \end{bmatrix} \times 4$ |
| Conv Block 3 | $\begin{bmatrix} 1 \times 3 \times 3 \text{ conv, } 256 \\ 3 \times 1 \times 1 \text{ conv, } 256 \end{bmatrix} \times 6$ | $\begin{bmatrix} 3 \times 3 \times 3 \text{ conv, } 256 \\ 3 \times 3 \times 3 \text{ conv, } 256 \end{bmatrix} \times 6$ | $\begin{bmatrix} 1 \times 1 \times 1 \text{ conv, } 216 \\ 3 \times 3 \times 3 \text{ conv, } 216 \\ 1 \times 1 \times 1 \text{ conv, } 96 \end{bmatrix} \times 6$ |
| Conv Block 4 | $\begin{bmatrix} 1 \times 3 \times 3 \text{ conv, } 512 \\ 3 \times 1 \times 1 \text{ conv, } 512 \end{bmatrix} \times 3$ | $\begin{bmatrix} 3 \times 3 \times 3 \text{ conv, } 512 \\ 3 \times 3 \times 3 \text{ conv, } 512 \end{bmatrix} \times 3$ | $\begin{bmatrix} 1 \times 1 \times 1 \text{ conv, } 432 \\ 3 \times 3 \times 3 \text{ conv, } 432 \\ 1 \times 1 \times 1 \text{ conv, } 192 \end{bmatrix} \times 3$ |
| FC | Average Pooling<br>Linear, 3<br>Softmax | Average Pooling<br>Linear, 3<br>Softmax | Average Pooling<br>Linear, 2048<br>Softmax |

*notes:* convolution layers include batch normalization, ReLU activation, and residual connections; filter sizes and output channels are indicated

used 20-frame input clips resized to $256 \times 256$ pixels. Data augmentation included random frame-wise color jittering and horizontal flipping. Transformer-based models were trained with the AdamW optimizer at a learning rate of 0.0001 under the same input configuration. Architectural specifications of the evaluated 3D-CNN backbones are detailed in Table 7.

## Online validation details

**Correction procedure** During online validation, the equipment FSM occasionally entered the `Error` state due to misdetected activity sequences. Revisions were performed manually following a structured correction procedure: (1) event logs and FSM states were reviewed, (2) the specific misdetection was identified, (3) the appropriate correction was applied, and (4) a `Recover` event was issued to return the FSM to a valid state (Idle or Holding).

**Detected events per crane** Table 8 summarizes the number of detected events for each crane across the three operational periods.

**Detected events per workpiece** Table 9 reports the event counts for each workpiece across the three operational periods, from `Incoming` registration to `Outgoing` dispatch.

**Extended EDA curves** Figures 13 and 14 present the event detection accuracy (EDA) curves separately for each crane and each workpiece, complementing the aggregate results reported in Section 4.3.

**Table 8** Detected counts of events for each crane across operational periods

| Period | Crane | Event | | | | |
|---|---|---|---|---|---|---|
| | ID | Grasp | Release | None | Fail | Recover Idle/Hold |
| 1 | $Crane_1$ | 53 | 52 | 62 | 11 | 6 / 5 |
| | $Crane_2$ | 4 | 3 | 3 | 1 | 1 / 0 |
| 2 | $Crane_1$ | 50 | 49 | 76 | 17 | 9 / 8 |
| | $Crane_2$ | 7 | 6 | 6 | 1 | 1 / 0 |
| 3 | $Crane_1$ | 32 | 32 | 38 | 4 | 2 / 2 |
| | $Crane_2$ | 6 | 6 | 10 | 0 | 0 / 0 |

**Table 9** Detected counts of events for each workpiece across operational periods

| Period | Workpiece | Event | | | |
|---|---|---|---|---|---|
| | ID | Incoming | Grasp | Release | Outgoing |
| 1 | WP11 | 1 | 7 | 7 | 1 |
| | WP12 | 1 | 7 | 7 | 1 |
| | WP13 | 1 | 9 | 9 | 1 |
| | WP14 | 1 | 9 | 9 | 1 |
| | WP15 | 1 | 8 | 8 | 1 |
| | WP16 | 1 | 8 | 8 | 1 |
| | WP17 | 1 | 7 | 7 | 1 |
| | WP18 | 1 | 7 | 7 | 1 |
| 2 | WP21 | 1 | 7 | 7 | 1 |
| | WP22 | 1 | 10 | 10 | 1 |
| | WP23 | 1 | 8 | 8 | 1 |
| | WP24 | 1 | 7 | 7 | 1 |
| | WP25 | 1 | 7 | 7 | 1 |
| | WP26 | 1 | 9 | 9 | 1 |
| | WP27 | 1 | 8 | 8 | 1 |
| | WP28 | 1 | 9 | 9 | 1 |
| 3 | WP31 | 1 | 7 | 7 | 1 |
| | WP32 | 1 | 5 | 5 | 1 |
| | WP33 | 1 | 6 | 6 | 1 |
| | WP34 | 1 | 6 | 6 | 1 |
| | WP35 | 1 | 5 | 5 | 1 |
| | WP36 | 1 | 5 | 5 | 1 |
| | WP37 | 1 | 6 | 6 | 1 |

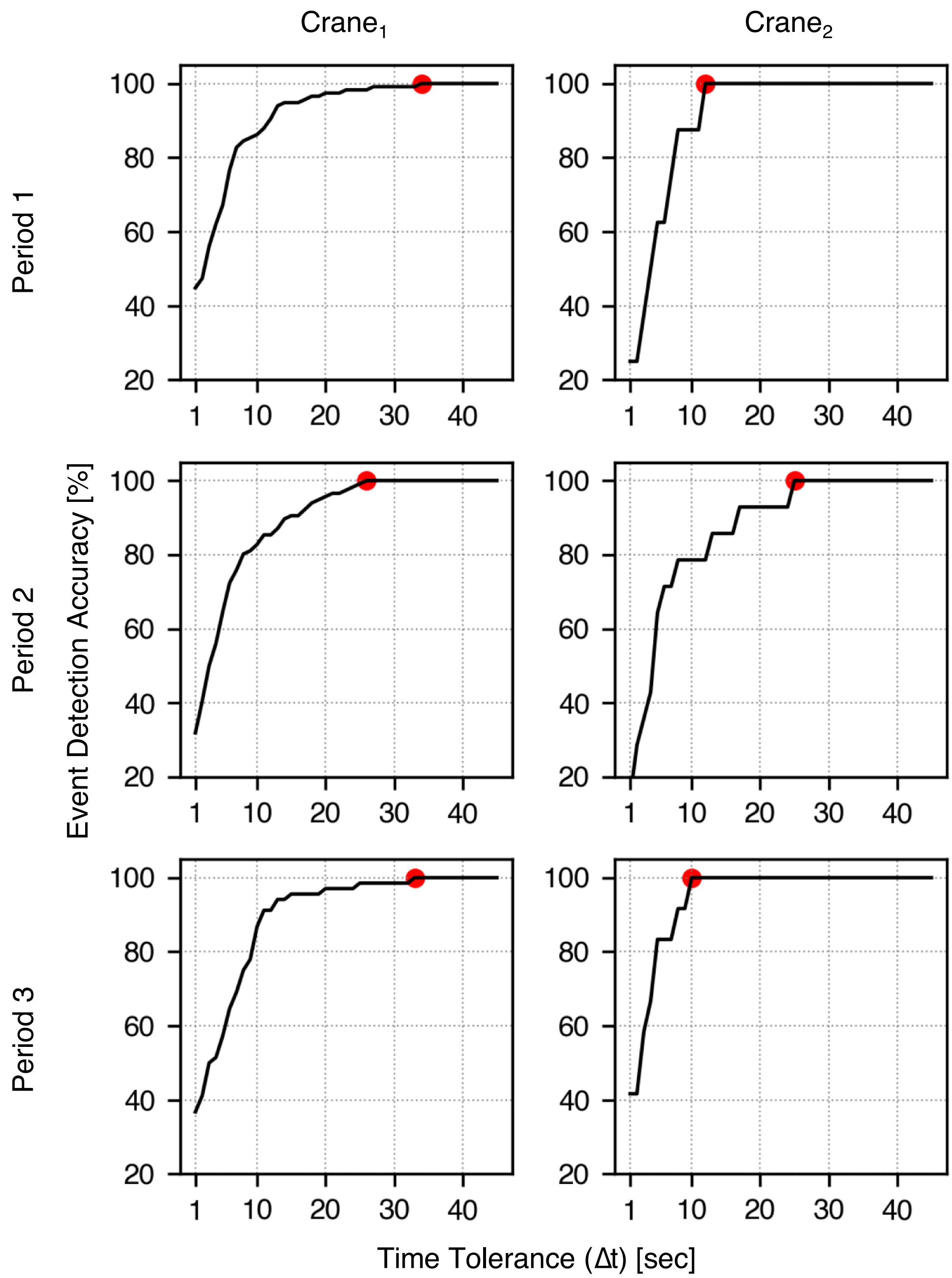

**Fig. 13** Event detection accuracy curves for $Crane_1$ and $Crane_2$ over time tolerances

$Crane_1$ $Crane_2$

Period 1

Period 2

Period 3

Event Detection Accuracy [%]

Time Tolerance (Δt) [sec]

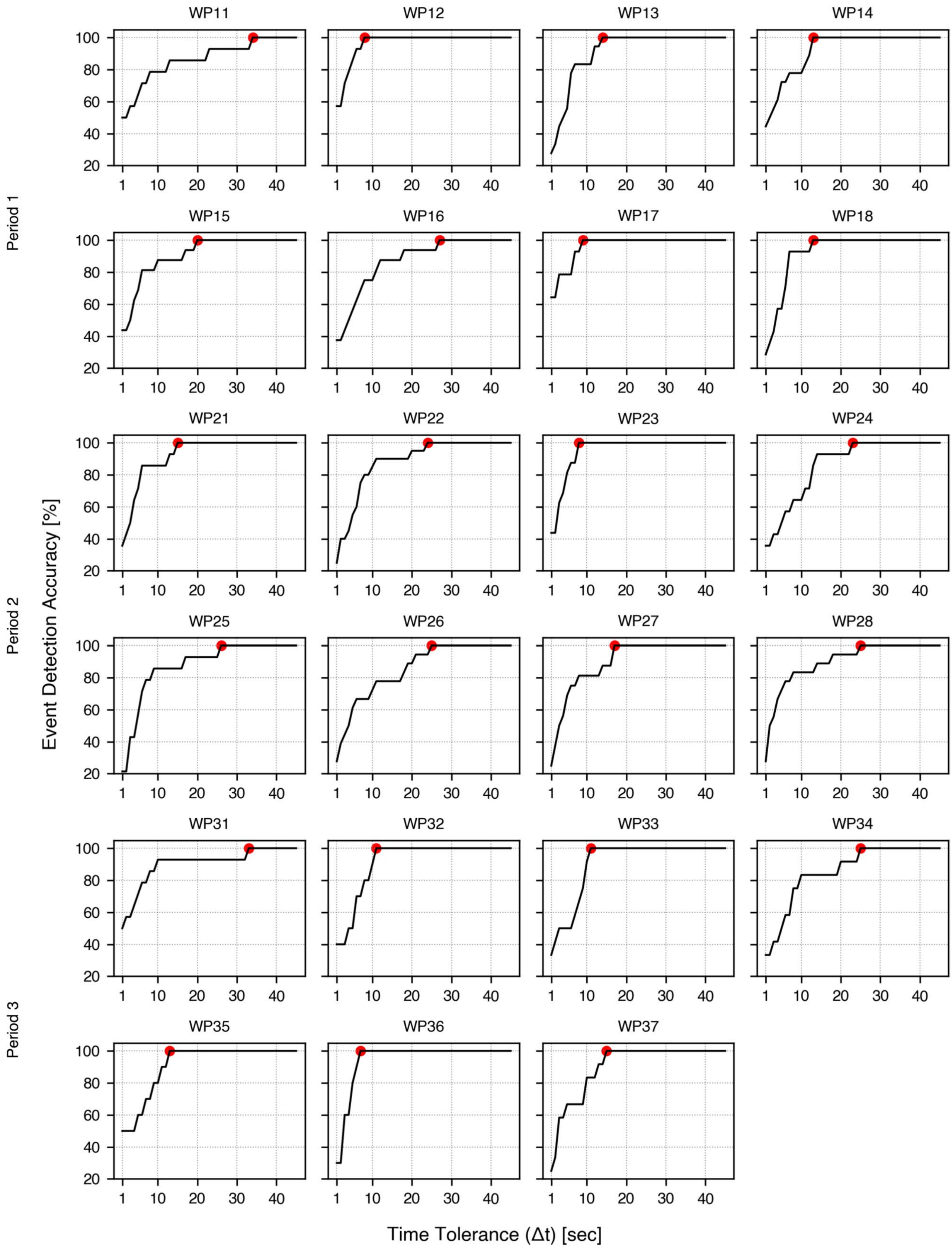

**Fig. 14** Detailed event detection accuracy curves for each workpiece over time tolerances

**Author Contributions** Dohyeon Kong: original draft writing, modeling, and experiments; Jaebong Cho: modeling and experiments; Hyunbo Cho: supervision and manuscript review.

**Funding** Open Access funding enabled and organized by Pohang University of Science and Technology (POSTECH). The authors declare that no funds, grants, or other support were received during the preparation of this manuscript.

## Declarations

**Competing Interests** The authors have no relevant financial or non-financial interests to disclose.

## References

1. Schino AD (2021) Open die forging process simulation: a simplified industrial approach based on artificial neural network. AIMS Mater Sci 8(5):685–697. https://doi.org/10.3934/matersci.2021041
2. Schuitemaker R, Xu X (2020) Product traceability in manufacturing: A technical review. Procedia CIRP 93:700–705. https://doi.org/10.1016/j.procir.2020.04.078. 53rd CIRP Conference on Manufacturing Systems 2020
3. Song L, Mohammed T, Stayshich D, Eldin N (2015) A cost effective material tracking and locating solution for material laydown yard. Procedia Eng 123:538–545. https://doi.org/10.1016/j.proeng.2015.10.106. Selected papers from Creative Construction Conference 2015
4. Zhao Q-J, Cao P, Tu D-W (2014) Toward intelligent manufacturing: label characters marking and recognition method for steel products with machine vision. Adv Manufac 2(1):3–12. https://doi.org/10.1007/s40436-014-0057-2
5. Liewald M, Karadogan C, Lindemann B, Jazdi N, Weyrich M (2018) On the tracking of individual workpieces in hot forging plants. CIRP J Manufac Sci Technol 22:116–120. https://doi.org/10.1016/j.cirpj.2018.04.002
6. Kang L-W, Chen Y-T, Jhong W-C, Hsu C-Y (2019) Deep learning-based identification of steel products. In: Pan J-S, Ito A, Tsai P-W, Jain LC (eds) Recent Advances in Intelligent Information Hiding and Multimedia Signal Processing. Springer, Cham, pp 315–323
7. Morar A, Moldoveanu A, Mocanu I, Moldoveanu F, Radoi IE, Asavei V, Gradinaru A, Butean A (2020) A comprehensive survey of indoor localization methods based on computer vision. Sensors 20(9). https://doi.org/10.3390/s20092641
8. Shim J-H, Cho Y-I (2015) A mobile robot localization using external surveillance cameras at indoor. Procedia Comput Sci 56:502–507. https://doi.org/10.1016/j.procs.2015.07.242
9. Dias Ja, Jorge PM (2015) People tracking with multi-camera system. In: Proceedings of the 9th international conference on distributed smart cameras. ICDSC '15, Association for Computing Machinery, New York, NY, USA, pp 181–186. https://doi.org/10.1145/2789116.2789141
10. Sun Y, Zhao K, Wang J, Li W, Bai G, Zhang N (2016) Device-free human localization using panoramic camera and indoor map. In: 2016 IEEE International conference on consumer electronics-China (ICCE-China), pp 1–5. https://doi.org/10.1109/ICCE-China.2016.7849743
11. Cosma A, Radoi IE, Radu V (2019) Camloc: Pedestrian location estimation through body pose estimation on smart cameras. In: 2019 International conference on indoor positioning and indoor navigation (IPIN), pp 1–8. https://doi.org/10.1109/IPIN.2019.8911770
12. Jain M, Nawhal M, Duppati S, Dechu S (2018) Mobiceil: cost-free indoor localizer for office buildings. In: Proceedings of the 20th international conference on human-computer interaction with mobile devices and services. MobileHCI '18. Association for Computing Machinery, New York, NY, USA. https://doi.org/10.1145/3229434.3229447
13. Pfitzner F, Braun A, Borrmann A (2024) From data to knowledge: Construction process analysis through continuous image capturing, object detection, and knowledge graph creation. Autom Construc 164:105451. https://doi.org/10.1016/j.autcon.2024.105451
14. Kim J, Chi S, Seo J (2018) Interaction analysis for vision-based activity identification of earthmoving excavators and dump trucks. Autom Construc 87:297–308. https://doi.org/10.1016/j.autcon.2017.12.016
15. Kim J, Chi S (2019) Action recognition of earthmoving excavators based on sequential pattern analysis of visual features and operation cycles. Autom Construc 104:255–264. https://doi.org/10.1016/j.autcon.2019.03.025
16. Roberts D, Golparvar-Fard M (2019) End-to-end vision-based detection, tracking and activity analysis of earthmoving equipment filmed at ground level. Autom Construc 105:102811. https://doi.org/10.1016/j.autcon.2019.04.006
17. Chen C, Zhu Z, Hammad A (2020) Automated excavators activity recognition and productivity analysis from construction site surveillance videos. Autom Construc 110:103045. https://doi.org/10.1016/j.autcon.2019.103045
18. Wang S, Yang L, Zhang Z, Zhao Y (2024) Keypoints-based heterogeneous graph convolutional networks for construction. Expert Syst Appl 237:121525. https://doi.org/10.1016/j.eswa.2023.121525
19. Li J, Wei P, Zheng N (2021) Nesting spatiotemporal attention networks for action recognition. Neurocomputing 459:338–348. https://doi.org/10.1016/j.neucom.2021.06.088
20. Vávra R, Filip J (2013) Registration of multi-view images of planar surfaces. In: Lee KM, Matsushita Y, Rehg JM, Hu Z (eds) Computer Vision - ACCV 2012. Springer, Berlin, Heidelberg, pp 497–509
21. Zhu X, Zhu Y, Wang H, Wen H, Yan Y, Liu P (2022) Skeleton sequence and rgb frame based multi-modality feature fusion network for action recognition. ACM Trans Multimed Comput Commun Appl 18(3). https://doi.org/10.1145/3491228
22. Wang Z, Bovik AC, Sheikh HR, Simoncelli EP (2004) Image quality assessment: from error visibility to structural similarity. IEEE Trans Image Process 13(4):600–612. https://doi.org/10.1109/TIP.2003.819861
23. Maji D, Nagori S, Mathew M, Poddar D (2022) YOLO-Pose: Enhancing YOLO for Multi Person Pose Estimation Using Object Keypoint Similarity Loss. arXiv:2204.06806
24. Bewley A, Ge Z, Ott L, Ramos F, Upcroft B (2016) Simple online and realtime tracking. In: 2016 IEEE International conference on

image processing (ICIP), pp 3464–3468. https://doi.org/10.1109/ICIP.2016.7533003
25. Tran D, Wang H, Torresani L, Ray J, LeCun Y, Paluri M (2018) A Closer Look at Spatiotemporal Convolutions for Action Recognition. arXiv:1711.11248
26. Hara K, Kataoka H, Satoh Y (2017) Learning Spatio-Temporal Features with 3D Residual Networks for Action Recognition. arXiv:1708.07632
27. Feichtenhofer C (2020) X3D: Expanding Architectures for Efficient Video Recognition. arXiv:2004.04730
28. Li Y, Wu C-Y, Fan H, Mangalam K, Xiong B, Malik J, Feichtenhofer C (2022) MViTv2: Improved Multiscale Vision Transformers for Classification and Detection. arXiv:2112.01526
29. Liu Z, Lin Y, Cao Y, Hu H, Wei Y, Zhang Z, Lin S, Guo B (2021) Swin Transformer: Hierarchical Vision Transformer using Shifted Windows. arXiv:2103.14030
30. He K, Zhang X, Ren S, Sun J (2015) Deep Residual Learning for Image Recognition. arXiv:1512.03385
31. Khanam R, Hussain M (2024) YOLOv11: An Overview of the Key Architectural Enhancements. arXiv:2410.17725